%% file: main.tex
\documentclass{article}

\usepackage{microtype}
\usepackage{graphicx}
\usepackage{subfigure}
\usepackage{booktabs} 
\usepackage{siunitx}

\usepackage{pgf}
\usepackage{pgfplots}
\pgfplotsset{compat=1.9}
\usepackage{tikz}
\usetikzlibrary{decorations.pathreplacing}
\usetikzlibrary{decorations.markings}
\usetikzlibrary{patterns}
\usepackage[outline]{contour}
\contourlength{1.2pt}
\usetikzlibrary{plotmarks}
\usetikzlibrary{spy}
\usetikzlibrary{calc}
\usetikzlibrary{shapes,plotmarks,positioning,fit,chains,tikzmark,arrows,hobby,arrows.meta}
\usetikzlibrary{decorations.pathreplacing,calligraphy}
\usepgfplotslibrary{groupplots}

\usepackage{hyperref}
\usepackage{empheq}

\usepackage[accepted]{icml2025}

\usepackage{amsmath}
\usepackage{amssymb}
\usepackage{mathtools}
\usepackage{amsthm}
\usepackage[normalem]{ulem}

\usepackage{enumitem}
\setlist[itemize]{noitemsep, topsep=-0.3em}

\icmltitlerunning{Theoretical Performance Guarantees for Partial Domain Adaptation via Partial Optimal Transport}

\usepackage{IEEEtrantools}
\usepackage{vmr-symbols-vecbold}
\usepackage{standard-macros}
\newcommand{\update}[1]{{#1}}

\input{tablesandfigures.tex}

\newcommand{\B}{\boldsymbol}
\newcommand{\R}{\mathbb{R}}

\newcommand{\Z}{\mathcal{Z}}
\newcommand{\X}{\mathcal{X}}
\newcommand{\Y}{\mathcal{Y}}
\newcommand{\E}{\mathbb{E}}
\newcommand{\I}{\mathbb{I}}
\let\W\undefined
\newcommand{\W}{\mathcal{W}}
\renewcommand{\P}{\mathbb{P}}
\newcommand{\mapping}[3]{#1:#2\rightarrow #3}

\usepackage{xspace}
\newcommand{\algoname}{\textsc{WARMPOT}\xspace}
\newcommand{\algonamelong}{\emph{weighted and regularized minimizer via partial optimal transport}\xspace}
\newcommand{\algonamearpm}{ARPM+our-weights\xspace}
\newcommand{\ie}{\emph{i.e.}}
\newcommand{\tildepi}{{\hat{p}_i}}
\newcommand{\tildeqj}{{\hat{q}_j}}
\newcommand{\bp}{{\hat{\B{p}}}}
\newcommand{\bpedge}[1]{{\hat{p}_{#1}}}
\newcommand{\bq}{{\hat{\B{q}}}}
\newcommand{\bqedge}[1]{{\hat{q}_{#1}}}
\newcommand{\Qztildeq}{{Q_{\B{\tilde{z}}}^{\hat{\B q}}}}
\newcommand{\primeg}{{g^{\prime}}}
\newcommand{\primew}{{w^{\prime}}}
\newcommand{\hatPistar}{{\hat{\Pi}^{\star}}}
\newcommand{\hatPistarij}{{\hat{\Pi}^{\star}_{ij}}}
\newcommand{\hatLf}{{\hat{L}_f}}
\newcommand{\hatLF}{{\hat{L}_F}}

\newcommand{\PWa}{\mathbb{PW}_\alpha}

\newcommand{\Psf}{P_{\B s}^f}
\newcommand{\Qtf}{Q_{\B t}^f}
\newcommand{\Pzf}{P_{\B z}^f}
\newcommand{\bPzf}{\dfrac1\beta P_{\B z}^f}
\newcommand{\Qtw}{Q_{\B t}^w}
\newcommand{\PSF}{P_{\B S}^F}
\newcommand{\QTF}{Q_{\B T}^F}
\newcommand{\PZF}{P_{\B Z}^F}

\newcommand{\QTW}{Q_{\B T}^W}
\newcommand{\Qz}{Q_{\B{\tilde{z}}}}
\newcommand{\Qzq}{Q_{\B{\tilde{z}}}^{\B q}}
\newcommand{\righto}{\right}
\newcommand{\tco}[1]{#1}
\renewcommand{\vecz}{\B z} 
\newcommand{\tBZ}{\tilde{\B Z}}
\newcommand{\Ltz}{L_{\B{\tilde{z}}}}
\newcommand{\LtZ}{L_{\tBZ}}
\newcommand{\TV}{\mathrm{TV}}
\newcommand{\prior}{Q_W}
\newcommand{\posterior}{P_{W\vert \B Z, \B T}}
\newcommand{\targpoploss}{L_{Q_{\tilde Z} }}
\newcommand{\KL}{D_{\rm KL}}
\renewcommand{\relent}[2]{\KL(#1\,||\,#2)}
\newcommand{\Uwj}{U^{(w)}_j}
\newcommand{\alphamax}{\alpha_{\textrm{max}}}
\renewcommand{\epsilon}{\varepsilon}

\usepackage[capitalize]{cleveref}
\theoremstyle{plain}
\newtheorem{theorem}{Theorem}[section]

\newtheorem{lemma}[theorem]{Lemma}
\newtheorem{corollary}[theorem]{Corollary}
\theoremstyle{definition}
\newtheorem{definition}[theorem]{Definition}

\theoremstyle{remark}

\crefname{theorem}{Theorem}{Theorems}
\crefname{lemma}{Lemma}{Lemmas}
\crefname{remark}{Remark}{Remarks}
\crefname{corollary}{Corollary}{Corollaries}
\crefname{definition}{Definition}{Definitions}
\crefname{proposition}{Proposition}{Propositions}

\begin{document}
\renewcommand{\arraystretch}{1.2}
\twocolumn[
	\icmltitle{Theoretical Performance Guarantees for Partial Domain Adaptation\\ via Partial Optimal Transport}



	\icmlsetsymbol{equal}{*}

	\begin{icmlauthorlist}
		\icmlauthor{Jayadev Naram}{chalmers}
		\icmlauthor{Fredrik Hellström}{ucl}
		\icmlauthor{Ziming Wang}{chalmers}
		\icmlauthor{Rebecka Jörnsten}{chalmers}
		\icmlauthor{Giuseppe Durisi}{chalmers}
	\end{icmlauthorlist}

	\icmlaffiliation{chalmers}{Chalmers University of Technology, Gothenburg, Sweden}
	\icmlaffiliation{ucl}{University College London, London, England}

	\icmlcorrespondingauthor{Jayadev Naram}{jayadev@chalmers.se}

	\icmlkeywords{Partial Domain Adaptation, Optimal Transport, Generalization Bounds}

	\vskip 0.3in
]



\printAffiliationsAndNotice{}  

\begin{abstract}
	In many scenarios of practical interest, labeled data from a target distribution are scarce while labeled data from a related source distribution are abundant.
	One particular setting of interest arises when the target label space is a subset of the source label space, leading to the framework of partial domain adaptation (PDA).
	Typical approaches to PDA involve minimizing a domain alignment term and a weighted empirical loss on the source data, with the aim of transferring knowledge between domains.
	However, a theoretical basis for this procedure is lacking, and in particular, most existing weighting schemes are heuristic.
	In this work, we derive generalization bounds for the PDA problem based on partial optimal transport.
	These bounds corroborate the use of the partial Wasserstein distance as a domain alignment term, and lead to theoretically motivated explicit expressions for the empirical source loss weights.
	Inspired by these bounds, we devise a practical algorithm for PDA, termed \algoname.
	Through extensive numerical experiments, we show that \algoname is competitive with recent approaches, and that our proposed weights improve on existing schemes.
\end{abstract}

\section{Introduction}\label{sec:introduction}

In unsupervised domain adaptation, one has access to a set of labeled source data and a set of unlabeled target data, drawn from different but related distributions.
The aim is to use these data sets to learn a predictor that performs well on new instances from the target data distribution \citep{redko2020survey, farahani2021brief}.
In contemporary practice, it is common for classifiers that are pre-trained on large, diverse domains to be deployed on smaller domains, characterized by a smaller label space.
This motivates the framework of \emph{unsupervised partial domain adaptation} (PDA), wherein the target label space is a subset of the source label space \citep{cao2018partial}.

In PDA, the use of labeled source data from outlier classes during training typically has an adverse effect on test performance---a phenomenon termed \emph{negative transfer} \citep{cao2018partial}.
To alleviate this issue, several heuristic schemes to weight the source data during training have been proposed \citep{zhang2018importance, liang2020balanced, li2020deep,gu2021adversarial,Gu2024}.
However, a theoretical motivation is lacking for most weight selections.

In this work, we provide
theoretically motivated algorithms for PDA.
Specifically, we derive generalization bounds on the target population loss and devise training strategies that minimize them.
The bounds that we obtain involve a partial Wasserstein distance between the empirical feature distributions for the source and target data \citep{caffarelli2010free}.
This motivates the popular strategy of learning a feature map that aligns source and target features, followed by a predictor trained to classify the labeled source features.
We will refer to objectives that are minimized with this aim as \emph{domain alignment terms}.
While the standard Wasserstein distance has been widely used to analyze and design algorithms for domain adaptation \citep{courty-14a,courty-17a}, its partial counterpart is crucial to handle the existence of outliers~\cite{wang2024partial}.
Additionally, our bounds include weighted source training losses, where, in contrast to all results available in the literature, the weights arise constructively from the partial optimal transport problem associated with the partial Wasserstein distance.
This enables a principled weight selection for addressing negative transfer that, unlike the aforementioned heuristics, comes with a clear theoretical motivation in terms of transport plans.

Our bounds come in two flavors. First, similar to~\citet{shen2018wasserstein}, one bound
depends on the partial Wasserstein distance between the empirical distributions of the
source and target features.
Second, building upon the work of \citet{courty2017joint}, we obtain a bound that incorporates estimates of the unknown labels of the target samples.  This yields a partial transport problem  involving the joint empirical distribution of features and labels.
Each bound depends on two parameters, which intuitively correspond to the expected portion of outliers in the source and target data sets, respectively.
Concretely, the parameters determine the proportion of mass from each of the data sets that is accounted for in the transport problem.

Inspired by the bounds, we propose a novel algorithm for PDA, termed \algonamelong (\algoname), whose performance we
compare against state-of-the-art (SOTA) methods.

\paragraph{Contributions.}
In this work, we derive two new families of generalization bounds for PDA, and devise algorithms to minimize them.
In particular, our bounds:
\vspace{1mm}
\begin{itemize}
	\item explicitly depend on the learned feature map, motivating, for the first time in the context of PDA, the approach of partly aligning feature distributions;
    \vspace{1.2mm}
	\item yield explicit weights for source data points with a principled motivation, giving a theoretically grounded way to tackle negative transfer;
    \vspace{1.2mm}
	\item lead to algorithms that improve upon or are comparable to recent approaches to PDA;
    \vspace{1.2mm}
	\item give rise to weights that, when combined with the ARPM algorithm of \citet{Gu2024}, lead to SOTA results for the Office-Home data set.
\end{itemize}
\vspace{1mm}
Furthermore, compared to previous bounds obtained for the more restrictive domain-adaptation problem, our
proof techniques directly yield bounds that depend on the Wassserstein distance
between the empirical distributions of source and target features. In contrast, in
existing bounds, such as the ones proposed by~\citet{courty2017joint}
and~\citet{shen2018wasserstein}, the Wasserstein distance involves the actual source and target
distributions, and an additional concentration-of-measure step is required to express such
bounds
in terms of numerically computable empirical distributions.

\section{Related Work} 

The problem of unsupervised domain adaptation was first formalized and analyzed by \citet{ben2006analysis}.
They derived a generalization bound in terms of the so called $\mathcal H$-divergence, defined in terms of a hypothesis class $\mathcal H$. 
This divergence is bounded and can be efficiently estimated if the VC dimension of $\mathcal H$ is finite~\citep{ben2010theory}.
Motivated by this bound, \citet{ganin2016domain} proposed domain-adversarial training, wherein an approximation of the $\mathcal H$-divergence is used as a domain alignment term.
However, the worst-case nature of the VC dimension leads to bounds that are too weak to explain generalization in deep neural networks~\citep{nagarajan2019uniform, zhang-21a-rethinking}.

In order to exploit the geometry of the data distributions, \citet{courty-14a,courty-17a} proposed the use of optimal transport, and specifically the Wasserstein distance, for domain adaptation.
This approach was theoretically supported by \citet{redko2017theoretical} and \citet{shen2018wasserstein}, who derived bounds in terms of the Wasserstein distance between the source and target input distributions.
Notably, this alleviates the issues of uniform convergence associated with the $\mathcal H$-divergence.
Based on these bounds, \citet{shen2018wasserstein} proposed a domain alignment term, computed using a Wasserstein generative adversarial network (GAN) operating on empirical feature distributions~\citep{arjovsky2017wasserstein}.
\citet{courty2017joint} derived a bound in terms of the joint source and target instance-label distributions, where estimates appear in place of the unknown target labels.
\citet{damodaran2018deepjdot} drew inspiration from this bound to devise an algorithm using mini-batch optimal transport on the joint feature-label distribution.
However, it is worth noting that the bounds reviewed so far all depend on the instance distributions, and do not incorporate the learned feature map.
Hence, they do not fully motivate the typical practice of computing the Wasserstein distance between the empirical distributions of source and target features.

As mentioned, an important factor in solving the PDA problem is to appropriately weight the source data.
\citet{cao2018partial} proposed to use heuristic class-level weights based on the predictions on unlabeled target inputs.
\citet{li2020deep} used these class-level weights, along with a maximum mean discrepancy loss as the domain alignment term.
In addition to domain-adversarial training, \citet{zhang2018importance}
and \citet{cao2019learning} determined the weights based on how well a domain discriminator can predict whether a given input is from the source or target distribution.
\citet{liang2020balanced} proposed entropy-aware weights, along with an advanced alignment strategy, while \citet{gu2021adversarial,Gu2024} computed weights by minimizing the Wasserstein distance between a weighted source feature distribution and the target feature distribution. \citet{nguyen2022improving} used the joint distribution partial Wasserstein distance as the domain alignment term, but with uniformly weighted source data.
\citet{wang2024partial} proposed a partial Wasserstein-GAN, an extension of Wasserstein-GAN to PDA \citep{Wang2022PartialWA}.
Specifically, they considered a class-level weighting scheme with the 1-partial Wasserstein distance as domain alignment term.
\citet{li2022partial} derived a generalization bound for PDA in terms of {model smoothness}, and proposed to focus on smoothness rather than alignment to transfer knowledge between domains. \citet{fatras2021unbalanced} use a mini-batch joint distribution unbalanced optimal transport (UOT) cost \update{as domain alignment term}, along with uniformly weighted source data. \citet{chang2022unified} consider a more general setting where the proposed algorithm uses the marginals of the UOT transport plan to compute binary weights for target samples.

Generalization bounds for domain adaptation containing a weighted source loss term are reported \update{in the works of} \citet{tachet2020domain} and \citet{luo2024invariant}. \update{These bounds} do not use \update{a} generalized Wasserstein metric\update{, such as the partial Wasserstein metric,} as domain alignment term. Furthermore, the bounds rely on class-level weights defined in terms of unknown data distributions. These weights are then estimated using a method developed by \citet{lipton2018detecting}. However, these estimates are only guaranteed to be accurate if the so called generalized label shift assumption \citep{tachet2020domain} holds exactly, i.e., if the feature representation $Z = f(X)$ of the instance $X$ with label $Y$ is such that $P(Z|Y=y)=Q(Z|Y=y)$ for source distribution $P$ and target distribution $Q$. 

In this work, similar to \citet{nguyen2022improving, wang2024partial}, we use the partial Wasserstein distance as the domain alignment term.
However, unlike prior work, we derive generalization bounds that both corroborate the choice of domain alignment term and lead to a theoretically motivated weighting scheme for the empirical source loss.
Furthermore, unlike some of the empirical weighting methods used in the literature (e.g., class-level weights), our approach readily extends beyond classification settings.

\section{Theoretical Results}\label{sec:theory}
We now present our main theoretical results.
First, in \cref{sec:notation}, we formalize the problem setup and introduce the notation.
In \cref{sec:bound-on-emp-target-loss}, we obtain bounds on the empirical target loss for a fixed sample, which we leverage to obtain generalization bounds in \cref{sec:pac-bayes-bound}.
Some useful definitions and results are recalled in \cref{app:preliminaries}.
\subsection{Problem Setup and Proposed Approach}\label{sec:notation}
We next introduce the notation that we will use throughout this section.
We let $\Z = \X\times \Y$ be the source domain, where $\X\subseteq \R^d$ is the input space, equipped with the sigma-algebra $\Sigma_{X}$, and $\Y \subseteq \R$ is the source label space,
equipped with the sigma-algebra $\Sigma_{Y}$.
We consider a joint probability distribution $P_Z$ on $(\Z,\Sigma_{X} \otimes
	\Sigma_{Y})$, called the \emph{source distribution}.
Similarly, we let $\tilde{\Z} = \X\times\tilde{\Y}$ be the target domain, where $\tilde{\Y}$, equipped with
the sigma-algebra $\Sigma_{\tilde{Y}}$, is an unknown
subset of $\Y$. Furthermore, we introduce a second joint probability distribution
$Q_{\tilde{Z}}$
on $(\tilde{\Z},\Sigma_{X} \otimes \Sigma_{\tilde{Y}})$, termed the \emph{target distribution}.\footnote{In the remainder of the paper, we will not specify sigma-algebras if they are clear from the context.}

A hypothesis is a measurable function $w:\X \to \Y$.
	{In order to discuss feature alignment, we express each hypothesis as $w = g \circ f$, where $f$ is a feature extractor and $g$ is a classifier.}
Throughout the paper, we will for simplicity consider bounded loss functions $\mapping{\ell}{\Y\times \Y}{[0,1]}$.
We are interested in
determining a hypothesis $w$ within a suitably chosen hypothesis class $\W$ (to be introduced later) that minimizes the population target loss
\begin{equation}
	L_{Q_{\tilde{Z}}}(w) = \E_{(X,Y)\sim Q_{\tilde{Z}}} [\ell(w(X), Y)].
\end{equation}
To do so, in the PDA setup considered in this paper, we
have at our disposal a vector $\B z = (z_1,\ldots, z_{n_s})\in \Z^{n_s},$ with $z_i =(x_i, y_i)$,
of labeled source instances drawn independently from $P_Z$.
Additionally, we have a vector $\B t = (\tilde{x}_1, \ldots,\tilde{x}_{n_t})$
of unlabeled target instances drawn independently from $Q_X$, which is the marginal distribution on $\X$ induced by $Q_{\tilde Z}$.
Let $\B{\tilde{z}} = (\tilde{z}_1,\ldots, \tilde{z}_{n_t})\in \tilde{\Z}^{n_t},$ with $\tilde{z}_j = (\tilde{x}_j, \tilde{y}_j)$, be the corresponding
vector of labeled target instances. Since this vector is not available to the learner,
the learner cannot evaluate the empirical target loss
\begin{equation}\label{eq:emprirical_target_loss}
	\Ltz(w) = \dfrac{1}{n_t}\sum_{j=1}^{n_t} \ell(w(\tilde{x}_j), \tilde{y}_j).
\end{equation}
To overcome this issue, we will obtain an upper bound on this quantity in \cref{sec:bound-on-emp-target-loss}.
This bound contains a partial
Wasserstein distance term and a weighted version of the empirical source loss, in which the weights are a function of the
optimal coupling measure $\Pi^{\star}$ obtained when solving the optimization problem in the definition
of the partial Wasserstein distance.
This definition is provided below.
\begin{definition}[{\citealp[Eq.~(2.1)]{figalli2010optimal}, \citealp[Eq.~(1.8)]{caffarelli2010free}}]\label{def:pw}
	The partial Wasserstein distance {with parameter $\alpha$} between two measures\footnote{We do not require that the two measures are probability measures. In particular, in our setup we will have $P_{X}(\X)\geq 1$.} $P_X$ and $Q_{\tilde{X}}$ on $(\X,\Sigma_{X})$ is defined as
	\begin{equation}\label{eq:PW-generic}
		\PWa(P_X,Q_{\tilde{X}}) = \inf_{\Pi\in\Gamma_\alpha(P_X,Q_{\tilde{X}})} \int c(x,\tilde{x})\ \mathrm{d}\Pi(x,\tilde{x}),
	\end{equation}
	where $\mapping{c}{\X\times \X}{\R_+}$ is the so-called cost function (typically a metric) 
	and $\Gamma_\alpha(P_X,Q_{\tilde{X}})$ is the set of all nonnegative measures $\Pi$ on $\X \times \X$ for which $\Pi(\X\times \X) = \alpha$ and, for which, for all measurable sets $\mathcal{A}, \mathcal{B}\subseteq \X$, we have that $\Pi(\mathcal{A}\times \X) \le P_X(\mathcal{A})$ and $\Pi(\X\times \mathcal{B}) \le Q_{\tilde{X}}(\mathcal{B})$.
\end{definition}
For the case in which the measures of interest are discrete and
supported on $m$ and $n$ mass points, respectively, it is convenient to express $P_{X}$
and $Q_{\tilde{X}}$ as vectors of dimension $m$ and $n$, and the
(coupling) measure as an $m\times n$ nonnegative (coupling) matrix $\Pi$ with entries $\Pi_{ij}$. Then,
one can rewrite~\eqref{eq:PW-generic} as
\begin{equation}\label{eq:PW-discrete}
	\PWa(P_X,Q_{\tilde{X}}) =\! \min_{\Pi\in\Gamma_\alpha(P_X,Q_{\tilde{X}})}
	\sum_{i=1}^{m} \sum_{j=1}^{n} c(x_{i},\tilde{x}_{j}) \Pi_{ij}
\end{equation}
where $\Gamma_\alpha(P_X,Q_{\tilde{X}})$ is the set of nonnegative matrices that satisfy\footnote{Here and throughout the paper, vector inequalities should be
	interpreted entry-wise.}
$\Pi \B 1_{n} \leq P_{X}$, $ \Pi^{T} \B 1_{m} \leq Q_{\tilde{X}}$ and $\B 1^{T}_{m} \Pi \B 1_{n} =
	\alpha$.

\subsection{Bounds on the Empirical Target Loss}\label{sec:bound-on-emp-target-loss}
Next, we present our main theoretical results: two bounds on the empirical target
loss~\eqref{eq:emprirical_target_loss}.
Generally speaking, the bounds consist of four terms: (i) a
weighted average of the loss computed on the labeled source instances, (ii) a partial Wasserstein term,
(iii) a total variation term that allows us to make the bound explicit in the empirical target
loss,
and, similar to most theoretical bounds for domain adaptation available in the literature \citep{ben2006analysis,courty2017joint,shen2018wasserstein}, (iv) a non-computable term that dictates the difficulty of the PDA problem under consideration.

Drawing inspiration from~\citet{shen2018wasserstein}, we first present a bound in Theorem~\ref{thm:feature_ot_bound} in which the partial Wasserstein distance is between the empirical
distributions of the source and target features.
Then, drawing inspiration from~\citet{courty2017joint}, we extend it in
\cref{thm:joint_ot_bound} to the case in which the partial Wasserstein distance is between the
joint empirical distribution of source features and labels
and the joint empirical distribution of target features and predicted labels. 
While the feature-based approach can capture covariate shift, where only the marginal distributions on the input differ, a joint distribution-based approach is beneficial in the case of labeling distribution shift, \ie, when
the conditional distribution on labels given inputs also differ.
Both bounds are in terms of the $\PWa$ distance.
Partial domain adaptation is achieved by inflating the empirical source distribution by a parameter $1/\beta$, where
$0<\beta\leq 1$.
Intuitively, the parameter $\beta$ relates to the fraction of source instances we want to
associate to target instances, {whereas the parameter $\alpha$} corresponds to the fraction of
target instances we want to consider.
Hence, $\beta$ allows for partial domain adaptation,
while $\alpha$ can be used to avoid outliers in the target set.

\begin{theorem}[Feature-based bound]\label{thm:feature_ot_bound}
	Assume that the loss function $\ell$ is a metric on $\Y$ and consider the set $\W$ of
	hypotheses $w = g \circ f$ for which $g$ is $\gamma$-Lipschitz
	with respect to~$\ell$.
	Let $\Psf = \tfrac{1}{n_s} \sum_{i=1}^{n_s} \delta_{f(x_i)}$ and
	$\Qtf = \tfrac{1}{n_t} \sum_{j=1}^{n_t} \delta_{f(\tilde{x}_j)}$ be
	the empirical source and target feature distributions, respectively, with feature extractor $f$.
	Then, for all $w\in \W$ and all $\alpha,\beta\in (0, 1]$,
        \begin{multline}\label{eq:emp_bound_ot}
		\Ltz(w) \le \sum_{i=1}^{n_s} \frac{p_{i}}{\alpha} \ell(w(x_i), y_i)  +
		\frac{2}{\alpha} \PWa \lefto(\tfrac{1}{\beta}\Psf,\Qtf\righto) \\
		+ \dfrac{1}{2}\sum_{j=1}^{n_t}\left| \dfrac{1}{n_t} - \dfrac{q_j}{\alpha} \right| + 2L_f,
	\end{multline}
	where the cost function in $\PWa$ is $c(x,\tilde{x})= \gamma \| f(x)-
		f(\tilde{x}) \|$, and the weights $\{p_{i}\}$ and $\{q_{j}\}$ are given by
	\begin{equation}\label{eq:pi}
		p_{i} = (\Pi^{\star} \B 1_{n_{t}})_{i}, \quad i=1,\dots,n_{s}
	\end{equation}
	\begin{equation}\label{eq:qj}
		q_{j} = \bigl((\Pi^{\star})^{T} \B 1_{n_{s}}\bigr)_{j}, \quad j=1,\dots,n_{t}
	\end{equation}

	with $\Pi^{\star}$ being the optimal coupling matrix in the definition of $\PWa(\tfrac{1}{\beta}\Psf,\Qtf)$. Finally,
	\begin{equation}\label{eq:L_F}
		L_f = \underset{\primeg\in\mathcal G}{\min}\ \underset{\substack{z\in \B z \cup \tilde{\B{z}}\\z\,=\,(x,y)}}{\max}\ \ell(\primeg(f(x)), y),
	\end{equation}
	where
	$\mathcal G$ denotes the set of classifiers $\primeg$ associated to hypotheses in $\W$, and where, with an abuse of notation, $\vecz \cup \tilde{\vecz}$ denotes the set of all labeled source and target instances.
\end{theorem}
The proof of \cref{thm:feature_ot_bound} is provided in \cref{app:feature_ot_bound}.
The weights $p_i$ and $q_j$ arise constructively from the partial transport problem corresponding to $\PWa(\tfrac{1}{\beta}\Psf,\Qtf)$. The underlying intuition is that, when partially transporting the source data to the target data, the source samples that play the dominant role in the transportation plan should be the ones that are most similar to the target samples. Conversely, outliers are expected to be nearly ignored, leading to small values of the corresponding $p_i$ and $q_j$. The connection between this transportation view and the actual prediction problem of interest is formalized in the proof of the bound, detailed in Appendix~\ref{app:feature_ot_bound}.

Note that when $\alpha=1$, we have that $q_{j}=1/n_{t}$ and, hence, the third term on the
right-hand side of~\eqref{eq:emp_bound_ot} disappears.
Setting $\alpha=1$ is reasonable if we do not expect any outliers in the target set, and, hence, we want to consider all target instances equally. In contrast, $\alpha<1$ and $\beta=1$ may be suitable for the so called  \emph{open set} adaptation problem, where the target label space includes additional classes beyond the source label space \citep{panareda-busto17-10a}.

Next, we present an analogous bound, in which joint feature-label distributions are used in place of feature-only distributions in the partial Wasserstein term.

\begin{theorem}[Joint distribution-based bound]\label{thm:joint_ot_bound}
	Assume that the loss function $\ell$ is a metric on $\Y$ and $\zeta$-Lipschitz in each argument.
	Consider the set $\W$ of hypotheses $w = g \circ f$ for which $g$ is $\gamma$-Lipschitz with respect to the Euclidean distance.
	Let $\Pzf = \tfrac{1}{n_s} \sum_{i=1}^{n_s} \delta_{f(x_i), y_i}$ and $\Qtw = \tfrac{1}{n_t} \sum_{j=1}^{n_t} \delta_{f(\tilde{x}_j), w(\tilde x_j)}$ be the empirical joint source and estimated joint target distributions, respectively, for the hypothesis $w=g\circ f$.
	Then, for all $w\in \W$ and all $\alpha,\beta\in (0, 1]$,
	\begin{multline}\label{eq:joint_ot_bound}
		\Ltz(w) \le \sum_{i=1}^{n_s} \dfrac{\tildepi}{\alpha} \ell(w(x_i), y_i) + \frac1\alpha\PWa \lefto( \tfrac{1}{\beta}\Pzf, \Qtw \righto) \\
		+ \dfrac{1}{2}\sum_{j=1}^{n_t}\left| \dfrac{1}{n_t} - \dfrac{\tildeqj}{\alpha} \right| + \hatLf,
	\end{multline}
    where the underlying cost function for $\PWa$ is
	\begin{equation}\label{eq:courty_cost}
		c((x,y), (\tilde x, \tilde y)) = \zeta\gamma\vecnorm{f(x)- f(\tilde x)}+\ell(y, \tilde y),
	\end{equation}
	the weights $\{\tildepi\}$ and $\{\tildeqj\}$ are given by 
	\begin{equation}\label{eq:pi_tilde}
		\tildepi = (\hatPistar \B 1_{n_{t}})_{i}, \quad i=1,\dots,n_{s}
	\end{equation}
	\begin{equation}\label{eq:qj_tilde}
		\tildeqj = \bigl((\hatPistar)^{T} \B 1_{n_{s}}\bigr)_{j}, \quad j=1,\dots,n_{t}
	\end{equation}
	with $\hatPistar$ being the optimal coupling matrix in the definition of $\PWa \lefto( \tfrac{1}{\beta}\Pzf, \Qtw \righto)$, and 
	\begin{equation}\label{eq:joint_ot_bound_Lfb}
		\hatLf= \min_{\primeg \in \mathcal G} \bigg\{ \sum_{j=1}^{n_t} \dfrac{\tildeqj}{\alpha}
		\ell\lefto(\primeg(f(\tilde x_j)), \tilde y_j \righto) \bigg\} +\Xi
	\end{equation} 
    with $\Xi$ given in \eqref{eq:Xi} (see \cref{app:joint_ot_bound}).
\end{theorem}

The proof of \cref{thm:joint_ot_bound} is provided in \cref{app:joint_ot_bound}. 
Note that the cost in \eqref{eq:courty_cost} coincides with the one proposed in \citet{courty-17a}.
However, \update{it} is important to note that the values of the weights can differ between the two bounds.
Indeed, while the weights are given by similar expressions, the underlying optimal coupling matrix differs in general.
The same considerations on the role of the \update{weights detailed} after \cref{thm:feature_ot_bound} also apply to Theorem~\ref{thm:joint_ot_bound}. 
{Furthermore, while both $\hatLf$ in \eqref{eq:joint_ot_bound_Lfb} and $L_f $ in \eqref{eq:L_F} relate to the difficulty of the PDA problem under consideration, they are incomparable in general.}
Finally, some terms in~\eqref{eq:emp_bound_ot} have an extra factor of $2$ compared to the corresponding terms in~\eqref{eq:joint_ot_bound}, and the underlying cost function in the partial Wasserstein distance in \eqref{eq:joint_ot_bound} has an extra term.
This, in addition to the slight difference in assumptions on the loss, means that the two bounds given in~\eqref{eq:emp_bound_ot} and~\eqref{eq:joint_ot_bound} {cannot be compared in general beyond the discussion above.}

\subsection{PAC-Bayes Generalization Bounds}\label{sec:pac-bayes-bound}
In \cref{sec:bound-on-emp-target-loss}, we derived bounds on the empirical target loss for a fixed hypothesis in the PDA setting.
However, these results cannot be applied directly to learned hypotheses.
To proceed, we will use the \emph{PAC-Bayesian} approach \citep{mcallester1999pac, catoni2007pac}, which will allow us to obtain loss bounds for a learned hypothesis. These bounds hold with high probability over the choice of the training source and target samples.

While a wide array of PAC-Bayes generalization bounds are available \citep{alquier2021user, hellstrom-25a}, we restrict ourselves to the following one for simplicity.

\begin{lemma}\label{lemma:pac_bayes}
	Suppose that there exists a function $R:\mathcal W\times \mathcal Z^{n_s} \times \tilde{\mathcal Z}^{n_t}$ such that, for all $(w, \B z, \tilde{\B z})\in \mathcal W\times \mathcal Z^{n_s} \times \tilde{\mathcal Z}^{n_t}$,
	\begin{equation}\label{eq:assumed_R-bound}
		\Ltz(w) \leq R(w, \B z, \tilde{\B z}) .
	\end{equation}
	Let $\prior$ be a prior distribution on $\mathcal W$ and $\posterior$ a posterior distribution on $\mathcal W$ given the labeled source samples\footnote{We denote by $P_Z^{n_s}$ the $n_s$-fold product of $P_Z$. Similarly, $Q_{\tilde{Z}}^{n_t}$ stands for the $n_t$-fold product of $Q_{\tilde{Z}}$.} $\B Z\distas P_Z^{n_s}$ and the unlabeled target samples ${\B T}$.
	Here, ${\B T}$ is the projection on $\mathcal X^{n_t}$ of $\tilde{\B Z}\distas Q_{\tilde Z}^{n_t}$.
	Then, for every fixed $\lambda>0$ and $\delta \in (0,1)$, with probability at least $1-\delta$ over $(\B Z, \tilde{\B Z})$,
	\begin{multline}
		\E_{\posterior}[\targpoploss(W)] \leq \E_{\posterior}[R(W, \B Z, \tilde{\B Z})] + \frac{\lambda}{8n_t}\\
		+ \frac{\relent{\posterior}{\prior} + \log\frac1\delta }{\lambda}
	\end{multline}
	where $\KL$ denotes the Kullback-Leibler (KL) divergence.
\end{lemma}
We provide the proof of \cref{lemma:pac_bayes} in \cref{app:pac_bayes}.
Note that, in \cref{sec:bound-on-emp-target-loss}, we derived functions $R$ that can be combined with \cref{lemma:pac_bayes} to yield generalization bounds for PDA.
We present these bounds below, beginning with a feature-based bound, obtained by combining \cref{thm:feature_ot_bound} and \cref{lemma:pac_bayes}.

\begin{corollary}\label{cor:feature-pac-bayes-bound}
	Suppose that the assumptions of \cref{thm:feature_ot_bound} and \cref{lemma:pac_bayes} hold, and consider the same notation as used therein.
    Furthermore, denote the decomposition of the hypothesis $W$ as $W=G\circ F$.
	Then, for every choice of $\lambda>0$ and $\delta\in(0,1)$, with probability at least $1-\delta$ over $\B Z\distas P_Z^{n_s}$, $\tBZ \distas Q_{\tilde Z}^{n_t}$,
	\begin{multline}
		\!\!\!\!\E_{\posterior}[\targpoploss(W)] \!\leq\! B\!+\! \E_{\posterior}\!\bigg[ \! \sum_{i=1}^{n_s} \frac{p_{i}}{\alpha} \ell(W(X_i), Y_i) \\
		 +\! \frac{2}{\alpha} \PWa \lefto(\tfrac{1}{\beta}\PSF,\QTF\righto) \!+\! \dfrac{1}{2}\sum_{j=1}^{n_t}\left| \dfrac{1}{n_t} \!-\! \dfrac{q_j}{\alpha} \right| \!+\! 2L_F \bigg],
	\end{multline}
where $(X_i, Y_i)=Z_i$ are the entries of $\B Z$, we let $\B S$ denote the projection on $\X$ of $\B Z$, and we use the shorthand
\begin{equation}\label{def:B-shorthand}
    B = \frac{\lambda}{8n_t} + \frac{\relent{\posterior}{\prior} + \log\frac1\delta }{\lambda} .
\end{equation}
\end{corollary}

Next, we present a joint distribution-based bound, which follows by \cref{thm:joint_ot_bound} and \cref{lemma:pac_bayes}.

\begin{corollary}
	Suppose that the assumptions of \cref{thm:joint_ot_bound} and \cref{lemma:pac_bayes} hold, and consider the same notation as used therein and in \cref{cor:feature-pac-bayes-bound}.
	Then, for every choice of $\lambda>0$ and $\delta\in(0,1)$, with probability at least $1-\delta$ over $\B Z\distas P_Z^{n_s}$, $\tBZ \distas Q_{\tilde Z}^{n_t}$,
    \begin{multline}
		\!\!\!\!\E_{\posterior}[\targpoploss(W)] \!\leq\! B\!+\! \E_{\posterior}\!\bigg[ \! \sum_{i=1}^{n_s} \frac{\tildepi}{\alpha} \ell(W(X_i), Y_i) \\
		 +\! \frac1\alpha\PWa \lefto( \tfrac{1}{\beta}\PZF, \QTW \righto) \!+\! \dfrac{1}{2}\sum_{j=1}^{n_t}\left| \dfrac{1}{n_t} \!-\! \dfrac{\tildeqj}{\alpha} \right| \!+\! \hatLF \bigg] .
    \end{multline}
\end{corollary}

The proof techniques that we use to derive these generalization bounds yield several key advantages compared to prior results obtained in the more restrictive context of domain adaptation by~\citet{shen2018wasserstein} and~\citet{courty2017joint}.
First, when relating the loss in the target domain to the loss in the source domain in \cref{sec:bound-on-emp-target-loss}, we work directly with the empirical measures.
Consequently, the partial Wasserstein distances in our results are fully empirical.
In contrast, in earlier derivations for the case of domain adaptation, the Wasserstein distance is between population measures, which need to be related to their empirical counterparts via a concentration-of-measure step.
This step adds an additional term to the bound, the need for which is obviated by our approach.
Furthermore, while most existing domain-adaptation algorithms compute domain alignment terms on the basis of learned feature distributions, the underlying generalization bounds available in the literature depend on the fixed input distribution.
In contrast, our bounds explicitly depend on the learned features, directly motivating the feature-based approaches most commonly used in practice.

\tableweights

\section{Algorithms for PDA: \algoname}\label{sec:algorithms}

Now, motivated by the bounds on the empirical target loss in \cref{sec:bound-on-emp-target-loss}, we propose a family of algorithms for the PDA problem.
Specifically, focusing on the first two computable terms of~\eqref{eq:joint_ot_bound} in \cref{thm:joint_ot_bound}, we consider the following optimization problem:
\begin{equation}\label{eq:pda_optimization}
	\min_{w} \left\{ \sum_{i=1}^{n_s} \tildepi \ell(w(x_i), y_i) + \PWa \lefto( \tfrac{1}{\beta}\Pzf, \Qtw \righto) \right\}.
\end{equation}
Here, the cost function in the definition of the partial Wasserstein distance is
$c((x,y), (\tilde x, \tilde y)) = \eta_1\|f(x)-f(\tilde x)\|+\eta_2\ell(y, \tilde y))$
and the weights $\tildepi$ are given by \eqref{eq:pi_tilde}.
By setting $\eta_2 = 0$, we observe that $\PWa ( \tfrac{1}{\beta}\Pzf, \Qtw )$ reduces to $\PWa (\tfrac{1}{\beta}\Psf,\Qtf )$, which is the domain alignment term appearing in~\eqref{eq:emp_bound_ot} in \cref{thm:feature_ot_bound}.
Thus, the optimization problem in~\eqref{eq:pda_optimization} allows us to draw inspiration from both Theorem~\ref{thm:feature_ot_bound} and Theorem~\ref{thm:joint_ot_bound}.

The parameters $\eta_1$ and $\eta_2$ {determine the impact of} the inter-feature and inter-label distances respectively, which act as regularizers, while $\alpha$ and $\beta$ control the alignment between the source and target distributions.

We refer to an algorithm that minimizes \eqref{eq:pda_optimization} as \algonamelong (\algoname).
To clarify the role of the parameters and the relation to prior work, we discuss two extreme cases:
\begin{itemize}
	\item \algoname with $\beta = 1$. In this case, the partial Wasserstein term aligns an $\alpha$ fraction of the source distribution with an $\alpha$ fraction of the target distribution.
	      This is reminiscent of the MPOT algorithm, which uses a mini-batch approximation of $\PWa$ to solve the PDA problem \citep{nguyen2022improving}.
	      The key differences are that {(i): MPOT uses a different cost function, where the inter-feature cost is given by the squared distance $\|f(x)-f(\tilde x)\|^2$, and (ii): MPOT uses uniform weights for the source sample losses, rather than the $p_i$ of \algoname}.

	\item \algoname with $\alpha = 1$. Here, the source distribution is scaled so that its total mass is $1/\beta>1$.
	      Of this mass, $1$ unit is aligned with the entire target distribution, whose total mass is $1$. The PWAN algorithm \citep{wang2024partial} uses this alignment approach {along with the heuristic class-level weights of the BA$^3$US algorithm \citep{liang2020balanced}, detailed in \cref{sec:ablation_study}.} 
\end{itemize}

The proposed \algoname algorithm can then be interpreted as aligning an $\alpha$ fraction of the $\beta$-scaled source distribution with an $\alpha$ fraction of the target distribution.
The use of two parameters $\alpha$ and $\beta$ allows for asymmetry in this domain alignment process, {which is necessary if the proportion of outliers differs between the source and target data sets.}

\section{Experiments}

We now experimentally evaluate our proposed algorithm, \algoname, for PDA tasks.
Specifically, in \cref{sec:exp_setup}, we detail our experimental setup.
In \cref{sec:algo_implementation}, we discuss the implementation details of \algoname.
In \cref{sec:ablation_study}, we investigate the impact of our proposed weight choice.
Then, in \cref{sec:main_results},
we compare \algoname against existing PDA methods.
Finally, in \cref{sec:weight-visualization}, we \tco{visualize} the weights used in \algoname and provide insight into their role.
Additional details on the experiments are provided in \cref{app:experimental_details}.

\subsection{Setup}\label{sec:exp_setup}

In the experiments, we focus on the Office-Home data set \citep{office-home}, which consists of images of $65$ objects belonging to {$4$ different domains: \textbf{A}rt, \textbf{C}lipart, \textbf{P}roduct, and \textbf{R}eal-World.}
To construct a PDA task, we consider a source data set consisting of all labeled samples from one domain and a target data set consisting of unlabeled samples from the first $25$ classes of another domain.
We consider all $12$ possible combinations of source and target domains.
\tco{This PDA setup has been widely studied and was considered by, among others, \citet{nguyen2022improving,Wang2022PartialWA,Gu2024}.}

\subsection{Implementation of \algoname} \label{sec:algo_implementation}

We consider hypotheses $w = g\circ f$ consisting of a ResNet50 \citep{He_2016_CVPR} feature extractor $f$ pretrained on ImageNet \citep{russakovsky2015imagenet} and a fully connected network with a hidden layer of dimension $256$ as the classifier $g$.
	We solve \eqref{eq:pda_optimization} using stochastic gradient descent,
		and compute $\PWa ( \tfrac{1}{\beta}\Pzf, \Qtw )$ for each mini-batch following the method proposed by~\citet{nguyen2022improving}.
Throughout, we set $\ell$ to be the cross-entropy loss and
the parameters of the domain alignment term to be $(\alpha, \beta) = (0.8, 0.35)$.
The values of all other hyperparameters are provided in \cref{app:experimental_details}. \footnote{Open source Python implementation of \algoname: \url{https://github.com/JayD2106/WARMPOT}.} The results of a sensitivity analysis on $\alpha$ and $\beta$ are discussed in \cref{app:sensitivity_analysis}.

\tableOfficeHome

\subsection{Comparison with Existing Weighting Schemes}\label{sec:ablation_study}

In this section, we evaluate our weighting strategy.
Specifically, we compare our choice for the weights in~\eqref{eq:pi_tilde} with the following weighting schemes:
\begin{itemize}
	\item MPOT weighting strategy~\citep{nguyen2022improving}, with uniform weights $\tildepi = 1/n_s$.
	\item BA$^3$US weighting strategy~\citep{liang2020balanced}, where
	      \begin{equation}
		      \tildepi = \sum_{j=1}^{n_t} \I[w(\tilde x_j) = y_i]/ n_t.
	      \end{equation}
	      Here, $\tilde x_j$ is a target instance and $y_i$ is the label of the $i$th source instance $x_i$.
	\item ARPM weighting strategy \citep{Gu2024}, where
	      \begin{equation}\label{eq:ARPM-weight-minimization}
		      \min_{\substack{\bp\in \Delta\\\tildepi\ge 0}}\mathbb{W}_1\lefto( \sum_{i=1}^{n_s} \tildepi \delta_{f(x_i)}, \dfrac{1}{n_t}\sum_{j=1}^{n_t} \delta_{f(\tilde{x}_j)} \righto).
	      \end{equation}
	      Here, $\bp = (\bpedge{1}, \ldots, \bpedge{n_s})$ and $\Delta$ is given by
            \begin{equation}\label{eq:ARPM-constraint-set}
		      \hspace{-0.2em}\Delta=\Bigg\{\bp :  \sum_{i=1}^{n_s} \tildepi = 1, \sum_{i=1}^{n_s} \left(\tildepi - \frac{1}{n_s}\right)^{\!2}\!\! < \frac{\rho}{n_s}\Bigg\}
            \end{equation}
          with $\rho$ being a hyperparameter.
\end{itemize}

The results are presented in \cref{table:ablation_office}. 
Note that, in our weighting strategy, the weights are computed for each mini-batch at every iteration. On the contrary, in BA$^3$US and ARPM, the weights are computed on the entire dataset but only every $n$ iterations.
We set this weight update interval to be $n = 500$ for BA$^3$US and ARPM as suggested in \citet{liang2020balanced} and \citet{Gu2024} respectively.
The experiments are repeated for 6 random seeds, and we report the average and the standard deviation.
As seen in the table, \algoname results in better performance than MPOT and ARPM, and yield performance comparable to BA$^3$US. 
This illustrates that the weighting strategy suggested by the theoretical results reported in Section~\ref{sec:theory} is indeed effective.
Results from a similar experiment using the ImageNet $\rightarrow$ Caltech data set (see \cref{app:extra_numerical}) confirm these findings.
Interestingly, the ARPM weighting strategy can be seen as a variation of our weighting strategy.
We discuss this connection in further detail in \cref{app:ARPM-weights}.

\begin{figure*}[ht]
    \centering
    \resizebox{\textwidth}{!}{\input{weights_plot.tex}}
    \caption{The distribution of WARMPOT weights for the task P$\rightarrow$A.
			Most of the weights of the outlier classes are close to zero,
			suggesting that most of the outliers are successfully omitted when training the classifier.}
    \label{fig:mOT}
\end{figure*}
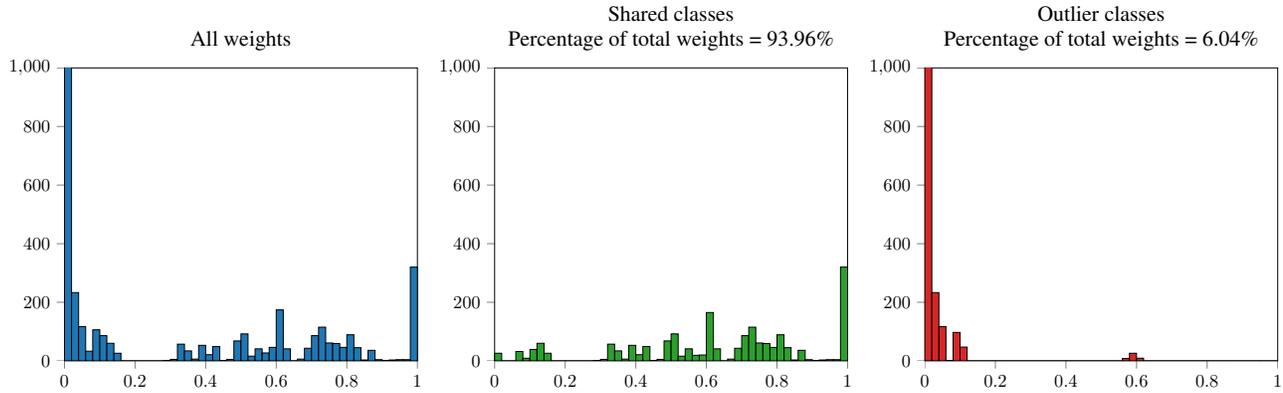

\subsection{Comparison with State of the Art}\label{sec:main_results}

Next, we compare the performance of \algoname with the performance of alternative algorithms proposed in the literature for the Office-Home data set. The results of our analysis are detailed in \cref{table:office_main}.\footnote{The test accuracy scores accompanied by standard deviation in \cref{table:office_main} are obtained by reproducing the results reported in the corresponding papers.}
First, we focus on comparisons to algorithms that, similar to \algoname, rely on a cost function of the form given in~\eqref{eq:pda_optimization}.
Specifically, we consider MPOT~\cite{nguyen2022improving} and PWAN~\cite{wang2024partial}.
As shown in the table, \algoname achieves higher average test accuracy compared to PWAN and similar to that of MPOT.

In the same table, we broaden the comparison to a wider range of algorithms.
As shown in the table, among the available algorithms, the best performance on Office-Home is achieved by ARPM~\cite{Gu2024}, which relies on several heuristic loss terms beyond the weighted source loss.
Motivated by our results in \cref{sec:ablation_study}, which indicate that the \algoname weighting strategy yields better performance than that of ARPM, we also consider an algorithm that is identical to ARPM except that it uses the \algoname weights.
We refer to this approach as \algonamearpm.
Interestingly,  \algonamearpm improves upon SOTA performance. This indicates that our theoretically motivated weighting strategy can lead to gains for PDA algorithms that involve a weighted source loss.
We detail the choice of hyperparameters used to obtain the test results of \algonamearpm in \cref{app:experimental_details}.

\subsection{The \algoname Weights}\label{sec:weight-visualization}

In order to illustrate the behavior of the \algoname weights, we focus on a single  task from the Office-Home data set, namely, the P$\rightarrow$A task.
In \cref{fig:mOT}, we present the weights $\tildepi$ in~\eqref{eq:pi_tilde} obtained by evaluating the transport matrix $\hatPistar$ achieving $\PWa ( \tfrac{1}{\beta}\Pzf, \Qtw )$ over the entire data set, at the end of the training process.
For illustrative purposes, we normalize the $\tildepi$ by $1/(\beta n_s)$ to obtain values in the interval $[0,1]$.
The left plot in \cref{fig:mOT} shows the distribution of all weights on source instances.
The middle and the right plots show the distribution of the normalized weights on the shared and the outlier class instances, respectively.
The weight proportion assigned to outlier source samples in \algoname is just $6.04\%$, even though $58.41\%$ of the source samples belong to outlier classes.
A total of $64.4\%$ outlier class instances have been assigned to the smallest bin, which helps in minimizing the effect of negative transfer.

\section{Conclusion}

In this work, we obtain generalization bounds for PDA tasks.
In particular, our bounds depend on a partial Wasserstein distance, and hence provide a theoretical motivation for using it as a domain alignment term.
While several existing algorithms in the literature take such an approach, a theoretical justification was previously missing.
Furthermore, our bounds constructively give rise to explicit source data weights, which can help alleviate negative transfer.
In contrast, prior work used heuristic weight choices, which were not directly motivated by theoretical considerations.

Inspired by our bounds, we propose the algorithm \algoname to minimize them.
Through numerical experiments, we demonstrate that \algoname is competitive with recent approaches to PDA.
Furthermore, we show that the performance of the SOTA algorithm ARPM is improved when its weighting scheme is replaced with that of \algoname.
This, along with an additional ablation study, corroborates the utility of our proposed weights.

It should be noted that an exact minimization of our bounds is prohibitively expensive from a computational standpoint, and hence, \algoname relies on some approximations.
Additional performance gains may be obtained by optimizing this implementation.
Furthermore, the SOTA algorithm ARPM includes additional loss terms that aim to reduce prediction uncertainty and improve robustness.
Such quantities are not explicitly present in our generalization bounds, but are studied by \citet[Thm.~1]{Gu2024}.
A promising direction for future research is to explore these aspects within our theoretical framework, potentially enabling more powerful algorithms.

\section*{Acknowledgements}

This work was partially supported by the Wallenberg AI, Autonomous Systems and Software Program (WASP) funded by the Knut and Alice Wallenberg Foundation. The computations were enabled by resources provided by the National Academic Infrastructure for Supercomputing in Sweden (NAISS), partially funded by the Swedish Research Council through grant agreement no. 2022-06725.

\section*{Impact Statement}

This paper presents work whose goal is to advance the field of 
machine learning. There are many potential societal consequences 
of our work, none which we feel must be specifically highlighted here.

\bibliography{references}
\bibliographystyle{icml2025}

\newpage
\appendix
\onecolumn

\section{Preliminaries}\label{app:preliminaries}
We recall some definitions that are used in the main text as well as in the appendices.
We will also establish a useful lemma.
\begin{definition}\label{def:metric}
	A function $\mapping{\rho}{\X\times\X}{\R_+}$ is called a metric on $\X$ if it is real-valued, finite, and nonnegative, and if
	for all $a, b, c \in \X$:
	\begin{enumerate}[label=(\roman*)]
		\item $\rho(a, b) = 0$ if and only if $a = b$,
		\item $\rho(a, b) = \rho(b, a)$\, (symmetry),
		\item $\rho(a, c) \le \rho(a, b) + \rho(b, c)$\, (triangle inequality).
	\end{enumerate}
\end{definition}
We shall also use the so-called reverse triangle inequality,
\begin{equation}\label{eq:reverse-triangle}
	|\rho(a,b) - \rho(a, c)| \le \rho(b, c)
\end{equation}
which can be readily obtained from the properties in Definition~\ref{def:metric}.

\begin{definition}\label{def:lipschitz}
	A function $\mapping{g}{\reals^{d}}{\Y}$ is $\gamma$-Lipschitz  
	with respect to a metric $\rho$ on $\Y$ if for
	all $t, t'\in \reals^{d}$
	\begin{equation}
		\rho(g(t), g(t')) \le \gamma \vecnorm{t-t'}.
	\end{equation}
\end{definition}

\begin{definition}\label{def:tv}
	The total variation distance between two discrete probability distributions $P$ and $Q$ on $\Z$ is defined as
	\begin{equation}
		\TV(P,Q) = \dfrac{1}{2} \sum_{z\in \Z} |P(z) - Q(z)|.
	\end{equation}
\end{definition}

\begin{lemma}\label{lemma:lipschitz-realistic}
	Assume that the loss function $\mapping{\ell}{\Y \times \Y}{\R_+}$ is a metric on $\Y$.
	Let $\W$ be the set of all hypotheses $w = g \circ f$ such that $g$ is $\gamma$-Lipschitz with respect to $\ell$. Let $L_f$ and $\vecz \cup \tilde{\vecz}$ be defined as in Theorem~\ref{thm:feature_ot_bound}.
	Then, for all pairs $z = (x, y)$ and $\tilde{z} = (\tilde{x}, \tilde{y})$ in $\vecz \cup \tilde{\vecz}$, we have
	\begin{equation}
		|\ell(w(x),y) - \ell(w(\tilde{x}), \tilde{y})| \le 2\gamma\|f(x) - f(\tilde{x})\| + 2L_f.
	\end{equation}
\end{lemma}
\begin{proof}
	Consider two arbitrary pairs $(x,y)$ and $(\tilde{x}, \tilde{y})$ in $\vecz \cup \tilde{\vecz}$.
	Furthermore, fix $f$ and let $g^{*}$ be a classifier achieving $L_{f}$
	in~\eqref{eq:L_F}.
	Then,
	\begin{IEEEeqnarray}{rCl}
		\ell(w(x),y) &=& \ell( g(f(x)), y ) \label{eq:lipschitz0} \\
		&\le& \ell( g(f(x)), g^*(f(x)) ) + \ell( g^*(f(x)), y ) \label{eq:lipschitz1} \\
		&\le& \ell( g(f(x)), g^*(f(x)) ) + L_f \label{eq:lipschitz2}  \\
		& \le & \ell( g(f(x)), g(f(\tilde{x}))  ) + \ell( g(f(\tilde{x})), g^*(f(x)) ) + L_f
		\label{eq:lipschitz3} \\
		& \le & \gamma \| f(x)-f(\tilde{x}) \| + \ell( g(f(\tilde{x})), g^*(f(x)) ) + L_f
		\label{eq:lipschitz4} \\
		& \le & \gamma \| f(x)-f(\tilde{x}) \| + \ell( g(f(\tilde{x})), g^*(f(\tilde{x})) )
		+ \ell( g^*(f(\tilde{x})), g^*(f(x)) ) + L_f \label{eq:lipschitz5} \\
		& \le & 2\gamma \| f(x)-f(\tilde{x}) \| + \ell( g(f(\tilde{x})), g^*(f(\tilde{x})) ) + L_f
		\label{eq:lipschitz6}  \\
		& \le & 2\gamma \| f(x)-f(\tilde{x}) \|  +  \ell( g(f(\tilde{x})), \tilde{y} ) + \ell( \tilde{y}, g^*(f(\tilde{x})) ) + L_{f}
		\label{eq:lipschitz6a} \\
		& \le & 2\gamma \| f(x)-f(\tilde{x}) \| + \ell(w(\tilde{x}),\tilde{y}) + 2 L_{f}. \label{eq:lipschitz7}
	\end{IEEEeqnarray}
	Here,~\eqref{eq:lipschitz0} follows because $w=g \circ f$; in~\eqref{eq:lipschitz1} we
	used the triangle inequality; in~\eqref{eq:lipschitz2} we used the max-min inequality \citep[Eq. (5.46)]{boyd04-a}
	as well as the definition of $L_{f}$;
	in~\eqref{eq:lipschitz3} we again used the triangle
	inequality; in~\eqref{eq:lipschitz4} we used that $g$ is $\gamma$-Lipschitz;
	in~\eqref{eq:lipschitz5} we used the triangle inequality; in~\eqref{eq:lipschitz6} we used
	that $g^*$ is $\gamma$-Lipschitz; in~\eqref{eq:lipschitz6a} we used the triangle
	inequality; and finally,~\eqref{eq:lipschitz7} follows again from the max-min inequality and
	the definition of $L_{f}$.
	Similarly, starting with $\ell(w(\tilde{x}),\tilde{y})$ and proceeding analogously,
	we conclude that
	\begin{equation}\label{eq:lipshitz_alt}
		\ell(w(\tilde{x}),\tilde{y}) \leq 2 \gamma \| f(x)-f(\tilde{x}) \| + \ell(w(x),y) + 2 L_{f}.
	\end{equation}
	Combining~\eqref{eq:lipschitz7} and~\eqref{eq:lipshitz_alt}, we obtain the desired result.
\end{proof}

\section{Proof of Theorem \ref{thm:feature_ot_bound}}\label{app:feature_ot_bound}
Note that there may be duplicate features in $\{f(x_i)\}_{i=1}^{n_s}$ and $\{f(\tilde x_i)\}_{i=1}^{n_t}$.
Hence, strictly speaking, $\Psf$ and $\Qtf$ are probability vectors whose dimensions are given by the number of distinct features, and multiplicities need to be accounted for.
However, in our proof, this yields the same result as if we treat the duplicate values as separate features with identical cost values.
Hence, for simplicity but without loss of generality, we assume that the features in both $\{f(x_i)\}_{i=1}^{n_s}$ and $\{f(\tilde x_i)\}_{i=1}^{n_t}$ are distinct.
This allows us to view $\Psf$ and $\Qtf$ as probability vectors of dimensions $n_s$ and $n_t$ respectively, where all entries of each vector are equal, \ie, $\Psf=[\frac1{n_s}, \dots, \frac{1}{n_s}]^T$ and $\Qtf=[\frac1{n_t}, \dots, \frac{1}{n_t}]^T$.

Now, define the ${n}_{s} \times {n}_{t}$ cost matrix $C$ with entries
$C_{ij}=\gamma \vecnorm{f(x_{i})- f(\tilde x_{j})}$.
We consider the partial Wasserstein distance  between
$\tfrac{1}{\beta}\Psf$ and $\Qtf$, which is given by (see the definition in~\eqref{eq:PW-discrete})
\begin{equation}\label{eq:pw1}
	\PWa\left(\tfrac{1}{\beta}\Psf,\Qtf\right) =
	\min_{\Pi\in \Gamma_\alpha\left(\tfrac{1}{\beta}\Psf,\Qtf\right)} \sum_{i=1}^{{n}_s} \sum_{j=1}^{{n}_t} C_{ij} \Pi_{ij},
\end{equation}
where $\Gamma_\alpha\lefto(\tfrac{1}{\beta}\Psf,\Qtf\righto) =
	\bigl\{\Pi\in \R^{{n}_s\times {n}_t}\;:\; \Pi \B 1_{{n}_t} \le \tfrac{1}{\beta}\Psf, \Pi^T \B 1_{{n}_s} \le \Qtf, \B 1^T_{{n}_s} \Pi \B 1_{{n}_t} = \alpha\bigr\}.$

Let $\Pi^{\star}\in \Gamma_\alpha\lefto(\frac{1}{\beta}\Psf,\Qtf\righto)$ attain the minimum in~\eqref{eq:pw1}.
Then
\begin{IEEEeqnarray}{rCl}
	2 \mathbb{PW}_{\alpha}\lefto
	( \frac{1}{\beta}\Psf,\Qtf\righto) & = & \sum_{i=1}^{{n}_s} \sum_{j=1}^{{n}_t} 2 C_{ij} \Pi^{\star}_{ij} \label{eq:boundB2}                                     \\
	& =& \sum_{i=1}^{{n}_s} \sum_{j=1}^{{n}_t} 2\gamma\|f(x_i)-f(\tilde{x}_j)\| \Pi^{\star}_{ij} \label{eq:boundB3}                                   \\
	& \ge& \sum_{i=1}^{{n}_s} \sum_{j=1}^{{n}_t} (|\ell(w(\tilde{x}_j), \tilde{y}_j) - \ell(w(x_i), y_i)| - 2L_f) \Pi^{\star}_{ij} \label{eq:boundB4}                                                   \\
	& \ge& \sum_{i=1}^{{n}_s} \sum_{j=1}^{{n}_t} (\ell(w(\tilde{x}_j), \tilde{y}_j) - \ell(w(x_i), y_i) - 2L_f) \Pi^{\star}_{ij} \label{eq:boundB5}                                          \\
	& =& \sum_{j=1}^{{n}_t} \ell(w(\tilde{x}_j),\tilde{y}_j) \sum_{i=1}^{{n}_s} \Pi^{\star}_{ij} - \sum_{i=1}^{{n}_s} \ell(w(x_i),y_i) \sum_{j=1}^{{n}_t} \Pi^{\star}_{ij} - 2L_f \sum_{i=1}^{{n}_s} \sum_{j=1}^{{n}_t} \Pi^{\star}_{ij} \label{eq:boundB6} \\
	& =& \sum_{j=1}^{{n}_t} \ell(w(\tilde{x}_j),\tilde{y}_j) ((\Pi^{\star})^T \B 1_{{n}_s})_j  - \sum_{i=1}^{{n}_s} \ell(w(x_i),y_i) (\Pi^{\star} \B 1_{{n}_t})_i - 2L_f \B 1^T_{{n}_s} \Pi^{\star} \B 1_{{n}_t} \label{eq:boundB7}                         \\
	& =& \sum_{j = 1}^{{n}_t} \ell(w(\tilde{x}_j),\tilde{y}_j){q}_j - \sum_{i=1}^{{n}_s} \ell(w(x_i),y_i) {p}_i - 2L_f\alpha. \label{eq:boundB10}
\end{IEEEeqnarray}
Here, in~\eqref{eq:boundB2} we used~\eqref{eq:pw1}; in~\eqref{eq:boundB3} we used the
definition of the cost function;~\eqref{eq:boundB4} follows from Lemma~\ref{lemma:lipschitz-realistic}; 
and finally, in~\eqref{eq:boundB10} we used that $\B 1^T_{{n}_s} \Pi^{\star} \B 1_{{n}_t} = \alpha$ by definition, and we set ${p}_{i}=(\Pi^{\star} \B 1_{{n}_t})_i$, $i=1,\dots, {n}_{s}$ and $q_{j} = \bigl((\Pi^{\star})^{T} \B 1_{n_{s}}\bigr)_{j}$, $j=1,\dots,{n}_{t}$.
The inequality just obtained can be rewritten as
\begin{equation}\label{eq:emp_alpha1}
	\sum_{j = 1}^{n_t} \ell(w(\tilde{x}_j),\tilde{y}_j){q}_j \le \sum_{i=1}^{n_s} \ell(w(x_i),y_i) {p}_i + 2 \mathbb{PW}_{\alpha}\lefto(\frac{1}{\beta}\Psf,\Qtf\righto) + 2\alpha L_f.
\end{equation}
Now, let $\B q = [{q}_1,\ldots, {q}_{n_t}]^T$ and define the following empirical distributions on $\B{\tilde{z}}$:
\begin{equation}\label{eq:q_dist}
	\Qz = \sum_{j=1}^{n_t} \dfrac{1}{n_t} \delta_{\tilde{z}_j},\qquad \Qzq = \sum_{j=1}^{n_t} \dfrac{{q}_j}{\alpha} \delta_{\tilde{z}_j}.
\end{equation}
Using that the loss function is supported on $[0,1]$, we perform the following change of
measure~\citep[Lemma 4]{ohnishi2021novel}:
\begin{equation}
	\E_{(X, Y)\sim \Qz} [\ell(w(X), Y)] \le \E_{(X, Y)\sim Q^{\B q}_{\B{\tilde{z}}}} [\ell(w(X), Y)] + \TV\lefto(\Qz, \Qzq\righto)
\end{equation}
where the total variation distance $\TV(\cdot,\cdot)$ was introduced in Definition~\ref{def:tv}.
This implies that
\begin{equation}
	\dfrac{1}{n_t}\sum_{j = 1}^{n_t} \ell(w(\tilde{x}_j),\tilde{y}_j)  \le \dfrac{1}{\alpha}\sum_{j = 1}^{n_t} \ell(w(\tilde{x}_j),\tilde{y}_j){q}_j + \TV\lefto(\Qz, \Qzq\righto)
\end{equation}
or, equivalently,
\begin{equation}
	\Ltz(w) \le \dfrac{1}{\alpha}\sum_{j = 1}^{n_t} \ell(w(\tilde{x}_j),\tilde{y}_j){q}_j + \TV\lefto(\Qz, \Qzq\righto). \label{eq:tv2}
\end{equation}
Finally, we note that
\begin{equation}\label{eq:tv_formula}
	\TV\lefto(\Qz, \Qzq\righto) = \dfrac{1}{2}\sum_{j=1}^{n_t}\left| \dfrac{1}{n_t} - \dfrac{{q}_j}{\alpha} \right|.
\end{equation}
We substitute~\eqref{eq:tv_formula} in~\eqref{eq:tv2} and~\eqref{eq:tv2} in~\eqref{eq:emp_alpha1} to get the desired result.

\section{Proof of \cref{thm:joint_ot_bound}}\label{app:joint_ot_bound}

As in \cref{app:feature_ot_bound}, we assume for simplicity that we can view $\Pzf$ and $\Qtw$ as probability vectors of dimensions $n_s$ and $n_t$ respectively, where all entries of each vector are equal, \ie, $\Pzf=[\frac1{n_s}, \dots, \frac{1}{n_s}]^T$ and $\Qtw=[\frac1{n_t}, \dots, \frac{1}{n_t}]^T$.

We now consider an ${n}_{s} \times {n}_{t}$ cost matrix $C$ with entries
$C_{ij}=\zeta\gamma\vecnorm{f(x_{i})- f(\tilde x_{j})}+\ell(y_i, w(\tilde x_j))$.
We consider the partial Wasserstein distance between
$\tfrac{1}{\beta}\Pzf$ and $\Qtw$, which is given by (see the definition in~\eqref{eq:PW-discrete})
\begin{equation}\label{eq:pwa-joint-pf}
	\PWa\lefto(\tfrac{1}{\beta}\Pzf,\Qtw\righto) =
	\min_{\Pi\in \Gamma_\alpha\lefto(\tfrac{1}{\beta}\Pzf,\Qtw\righto)} \sum_{i=1}^{{n}_s} \sum_{j=1}^{{n}_t} C_{ij} \Pi_{ij},
\end{equation}
where $\Gamma_\alpha\lefto(\tfrac{1}{\beta}\Pzf,\Qtw\righto) =
	\bigl\{\Pi\in \R^{{n}_s\times {n}_t}\;:\; \Pi \B 1_{{n}_t} \le \tfrac{1}{\beta}\Pzf, \Pi^T \B 1_{{n}_s} \le \Qtw, \B 1^T_{{n}_s} \Pi \B 1_{{n}_t} = \alpha\bigr\}.$
We define $\Qz$ and $\Qztildeq$ in the same way as in~\eqref{eq:q_dist}, where $\bq = [\bqedge{1}, \ldots, \bqedge{n_t}]^{T}$.
Then, given the feature map $f$, for every fixed hypothesis $\primew \in \mathcal W$ that can be decomposed as $\primew = \primeg \circ f$, we have
\begin{align}
	\alpha \Ltz(w) & \leq \alpha \TV(\Qz, \Qztildeq) + \sum_{j=1}^{n_t} \tildeqj \ell(w(\tilde x_j), \tilde y_j)                                                                  \label{eq:tv-trick-1} \\
	               & \leq \alpha \TV(\Qz, \Qztildeq) + \sum_{j=1}^{n_t} \tildeqj \left(\ell(w(\tilde x_j), \primew(\tilde x_j)) + \ell(\primew(\tilde x_j), \tilde y_j) \right)\label{eq:tv-trick-2}  \\
	               & =\! \alpha \TV(\Qz, \Qztildeq) \!+\! \sum_{i=1}^{n_s} \tildepi \ell(\primew(x_i), y_i) \!+\! \sum_{j=1}^{n_t} \!\tildeqj
	\ell(\primew(\tilde x_j), \tilde y_j) \!+\! \sum_{j=1}^{n_t} \!\tildeqj \ell(w(\tilde x_j), \primew(\tilde x_j)) \!-\! \sum_{i=1}^{n_s} \tildepi \ell(\primew(x_i), y_i) . \label{eq:joint-pf-decomposition}
\end{align}
Here,~\eqref{eq:tv-trick-1} follows from~\eqref{eq:tv2}; in~\eqref{eq:tv-trick-2} we used that the weights $\{\tildeqj\}$ are nonnegative as well as triangle inequality; to obtain~\eqref{eq:joint-pf-decomposition} we just summed and subtracted the term
$\sum_{i=1}^{n_s} \tildepi \ell(\primew(x_i), y_i)$.
We now focus on the last two terms of \eqref{eq:joint-pf-decomposition}.
Let $\hatPistar$ be the coupling matrix achieving $\PWa \lefto( \tfrac{1}{\beta}\Pzf, \Qtw \righto)$.
We have
\begin{align}\label{eq:joint-pf-pwa-bound-1}
	\sum_{j=1}^{n_t} \tildeqj \ell(w(\tilde x_j), \primew(\tilde x_j)) -\sum_{i=1}^{n_s} \tildepi \ell(\primew(x_i), y_i)
	 & = \sum_{j=1}^{n_t}  \ell(w(\tilde x_j), \primew(\tilde x_j)) \sum_{i=1}^{n_s} \hatPistarij -\sum_{i=1}^{n_s}  \ell(\primew(x_i), y_i) \sum_{j=1}^{n_t} \hatPistarij \\
	 & = \sum_{i=1}^{n_s} \sum_{j=1}^{n_t} \hatPistarij \left( \ell(w(\tilde x_j), \primew(\tilde x_j))   -  \ell(\primew(x_i), y_i) \right)                             \\
	 & \leq \sum_{i=1}^{n_s} \sum_{j=1}^{n_t} \hatPistarij \abs{ \ell(w(\tilde x_j), \primew(\tilde x_j))   - \ell(\primew(x_i), y_i) }
	\\ &\leq \sum_{i=1}^{n_s} \sum_{j=1}^{n_t} \hatPistarij \bigg[  \abs{ \ell(w(\tilde x_j), \primew(\tilde x_j))   -  \ell(w(\tilde x_j), \primew( x_i))   } \\
	 & \qquad\qquad\qquad\qquad  +  \abs{ \ell(w(\tilde x_j), \primew( x_i)) - \ell(\primew(x_i), y_i) } \bigg]  \nonumber
	\\ \label{eq:joint-pf-pwa-bound-Lip1} &\leq \sum_{i=1}^{n_s} \sum_{j=1}^{n_t} \hatPistarij  \left[ \zeta\abs{\primew(\tilde x_j) - \primew(x_i)} + \ell(w(\tilde x_j), y_i) \right] \\ \label{eq:joint-pf-pwa-bound-Lip2}
	 & \leq \sum_{i=1}^{n_s} \sum_{j=1}^{n_t} \hatPistarij  \left[ \zeta\gamma\left\|f(\tilde x_j) - f(x_i)\right\| + \ell(w(\tilde x_j), y_i) \right]                     \\
	 & = \PWa\lefto(\bPzf, \Qtw\righto) . \label{eq:joint-pf-pwa-bound-last}
\end{align}
Here, \eqref{eq:joint-pf-pwa-bound-1} follows from the definitions of $\tildepi$ in \eqref{eq:pi_tilde} and $\tildeqj$ in \eqref{eq:qj_tilde}; \eqref{eq:joint-pf-pwa-bound-Lip1} follows because the loss is $\zeta$-Lipschitz and because of the reverse triangle inequality~\eqref{eq:reverse-triangle}; and \eqref{eq:joint-pf-pwa-bound-Lip2} follows since $\primeg$ is $\gamma$-Lipschitz with respect to the Euclidean distance.

By substituting \eqref{eq:joint-pf-pwa-bound-last} into \eqref{eq:joint-pf-decomposition} and decomposing $\primew$ as $\primew=\primeg\circ f$, we obtain
\begin{equation}
	\Ltz(w) \leq \frac1\alpha\PWa\lefto(\bPzf, \Qtw\righto) + \TV(\Qz, \Qztildeq) + \sum_{i=1}^{n_s} \frac{\tildepi}{\alpha} \ell\lefto(\primeg(f(x_i)), y_i \righto) + \sum_{j=1}^{n_t} \frac{\tildeqj}{\alpha}
	\ell\lefto(\primeg(f(\tilde x_j)), \tilde y_j \righto). \label{eq:joint-pf-last-step}
\end{equation}

Next, we define
\begin{multline}\label{eq:Xi}
	\Xi = \min_{\primeg \in \mathcal G} \lefto\{\sum_{i=1}^{n_s} \frac{\tildepi}{\alpha} \ell\lefto(\primeg(f(x_i)), y_i \righto) + \sum_{j=1}^{n_t} \frac{\tildeqj}{\alpha}
	\ell\lefto(\primeg(f(\tilde x_j)), \tilde y_j \righto) \righto\} \\
    - \lefto( \min_{\primeg \in \mathcal G} \lefto\{\sum_{i=1}^{n_s} \frac{\tildepi}{\alpha} \ell\lefto(\primeg(f(x_i)), y_i \righto)\righto\} + \min_{\primeg \in \mathcal G} \lefto\{\sum_{j=1}^{n_t} \frac{\tildeqj}{\alpha}
	\ell\lefto(\primeg(f(\tilde x_j)), \tilde y_j \righto) \righto\}  \righto).
\end{multline}
We now minimize over $\primeg$ in the two summations of \eqref{eq:joint-pf-last-step}, and note that

\begin{multline}
	\min_{\primeg \in \mathcal G} \lefto\{\sum_{i=1}^{n_s} \frac{\tildepi}{\alpha} \ell\lefto(\primeg(f(x_i)), y_i \righto) + \sum_{j=1}^{n_t} \frac{\tildeqj}{\alpha}
	\ell\lefto(\primeg(f(\tilde x_j)), \tilde y_j \righto) \righto\} \\
    \leq \sum_{i=1}^{n_s} \frac{\tildepi}{\alpha} \ell\lefto( g(f(x_i)), y_i \righto) + \min_{\primeg \in \mathcal G} \lefto\{ \sum_{j=1}^{n_t} \frac{\tildeqj}{\alpha}
	\ell\lefto(\primeg(f(\tilde x_j)), \tilde y_j \righto) \righto\} + \Xi. \label{eq:joint-minimization}
\end{multline}

We obtain the desired result by recalling that $g(f(x_i))=w(x_i)$, by using the definition of $\hatLf$, and by writing out the explicit form of $\TV(\Qz, \Qztildeq)$, as per~\eqref{eq:tv_formula}.

\section{Proof of Lemma~\ref{lemma:pac_bayes}}\label{app:pac_bayes}

The proof follows that of \citet[Thm.~2.1]{alquier2021user}, with adjustments to account for the non-standard posterior and the use of $R(W, \B Z, \tilde{\B Z})$ in place of the empirical loss.

For a fixed $w\in \W$, we let $\Uwj = \targpoploss(w) - \ell(w(\tilde X_j), \tilde Y_j)$.
Note that the $\ell(w(\tilde X_j), \tilde Y_j)$ are supported on an interval of width $1$, are independent and identically distributed, and satisfy $\E[\ell(w(\tilde X_j), \tilde Y_j)]=\targpoploss(w)$.
Hence, we can apply Hoeffding's inequality \citep[Prop.~2.5]{wainwright-19a} to find that, for every $t>0$,
\begin{equation}
	\E_{\tilde{\B Z}\distas Q_{\tilde Z}^{n_t}}\lefto[ e^{t\sum_{j=1}^{n_t} \Uwj} \righto] \leq e^{\frac{t^2n_t}{8}}.
\end{equation}
Now, note that $\sum_{j=1}^{n_t} \Uwj = n_t(\targpoploss(w) -
	\LtZ(w))$. Hence, setting $t=\lambda/n_t$ for some $\lambda>0$, we have
\begin{equation}\label{eq:pb-pf-hoefflamb}
	\E_{\tilde{\B Z}\distas Q_{\tilde Z}^{n_t}}\lefto[ e^{
				\lambda\left( \targpoploss(w)-\LtZ(w) \right) } \righto] \leq e^{\frac{\lambda^2}{8n_t}}
\end{equation}
Now, it follows from the upper bound in~\eqref{eq:assumed_R-bound} that
\begin{equation}\label{eq:pb-pf-use-R-bound}
	\E_{\tilde{\B Z}\distas Q_{\tilde Z}^{n_t}}\lefto[ e^{
				\lambda\left( \targpoploss(w)-R(w,\B z, \tilde{\B Z}) \right) } \righto] \leq \E_{\tilde{\B Z}\distas Q_{\tilde Z}^{n_t}}\lefto[ e^{
				\lambda\left( \targpoploss(w)-\LtZ(w) \right) } \righto] .
\end{equation}
By combining \eqref{eq:pb-pf-hoefflamb} and \eqref{eq:pb-pf-use-R-bound} and averaging over $\B Z \distas P_Z^{n_s}$ and $W\distas Q_W$ we find that, collecting factors on the left-hand side,
\begin{equation}
	\E_{\B Z \distas P_Z^{n_s} , \tilde{\B Z}\distas Q_{\tilde Z}^{n_t}, W\distas Q_W }\lefto[ e^{
				\lambda\left( \targpoploss(W)-R(W,\B Z, \tilde{\B Z}) \right) - \frac{\lambda^2}{8n_t} } \righto] \leq 1.
\end{equation}
In the remainder of the proof, we will suppress the explicit distributions in the expectation notation when they are clear from context.
We now apply the Donsker-Varadhan variational formula \citep[Lemma 2.2]{alquier2021user} to conclude that,
for a given posterior $\posterior$,
\begin{equation}
	\E_{\B Z, \tBZ}\lefto[e^{\lambda \E_{W\distas\posterior}[\targpoploss(W)-R(W,\B Z, \tBZ)] -\relent{\posterior}{\prior} - \frac{\lambda^2}{8n_t} } \righto]  \leq 1.
\end{equation}
Next, by the Chernoff bound \citep[Eq. (2.5)]{wainwright-19a}, we have that for every $s>0$,
\begin{align}
	 & \P_{\B Z, \tBZ} \lefto[ \lambda \E_{W\distas\posterior}[\targpoploss(W)-R(W,\B Z, \tBZ)] -\relent{\posterior}{\prior} - \frac{\lambda^2}{8n_t} > s \righto] \nonumber  \\
	 & \leq \E_{\B Z, \tBZ}\lefto[e^{\lambda \E_{W\distas\posterior}[\targpoploss(W)-R(W,\B Z, \tBZ)] -\relent{\posterior}{\prior} - \frac{\lambda^2}{8n_t} } \righto] e^{-s} \\
	 & \leq e^{-s}.
\end{align}
Setting $s=\log(1/\delta)$, we thus conclude that with probability at most $\delta$ over $\B Z\distas P_Z^{n_s}$, $\tBZ \distas Q_{\tilde Z}^{n_t}$,
\begin{equation}
	\lambda \E_{W\distas\posterior}[\targpoploss(W)-R(W,\B Z, \tBZ)] -\relent{\posterior}{\prior} - \frac{\lambda^2}{8n_t} > \log\frac1\delta.
\end{equation}
We obtain the desired result by considering the complementary event and re-arranging terms.

\section{Additional Details on the Experiments} \label{app:experimental_details}

We compute the domain alignment term $\PWa \lefto( \tfrac{1}{\beta}\Pzf, \Qtw \righto)$ in \eqref{eq:pda_optimization} using the entropic partial Wasserstein solver from the POT library \citep{flamary2021pot}, with regularization constant $\epsilon = 7.0$ in all the experiments\update{, which is} selected to avoid numerical instabilities.
We set the maximum number of iterations to $\num{5000}$.
Following \citet{nguyen2022improving},
we linearly increase $\alpha$ from $0.01$ to $\alphamax$ for the first $\num{2500}$ iterations, and keep it constant for the last $\num{2500}$ iterations.
Through a parameter search, we obtained the following values for the hyperparameters: $\alphamax = 0.8, \eta_1 = 0.125, \eta_2 = 1.75, \beta = 0.35$.
We use a batch size of $65$, and set the learning rate of stochastic gradient descent to $0.001$. We used the same values for these hyperparameters in all our experiments.
For \algonamearpm, we set ${1}/{\beta}=3$, while all other hyperparameters are the same as those in ARPM \citep{Gu2024}.

\section{Additional Numerical Results}\label{app:extra_numerical}

In this section, we discuss the results obtained by repeating the experiments described in \cref{sec:ablation_study} on the ImageNet $\rightarrow$ Caltech dataset,
where ImageNet~\citep{russakovsky2015imagenet} consists of $1000$ classes and Caltech-256~\cite{griffin2007caltech} consists of $256$ classes. 

We first compare different weighting schemes.
Following \citet{Gu2024}, we set the weight update intervals of ARPM and BA$^3$US to 2000.
The experiment is repeated for 3 random seeds, and we report the average and \update{the} standard deviation.
The results are presented in~\cref{table:ablation_imagenet}.
\algoname results in better performance than MPOT and ARPM, and yields performance comparable to BA$^3$US.

Then we compare \algoname against alternative algorithms. We set $\alphamax = 0.08$, $\eta_1 = 0.92$, $\eta_2 = 5.47$, $\beta = 0.72$, $\epsilon = 5.59$ for \algoname. The results are shown in~\cref{table:imagenet_main}. Once more, we observe that \algonamearpm achieves better performance than ARPM, highlighting the effectiveness of the \algoname weights.

\tableweightsimagenet

\tableImageNet

\section{Sensitivity Analysis for Alignment Parameters}\label{app:sensitivity_analysis}

In order to assess the impact on performance of the hyperparameters in the domain alignment term, we conduct a sensitivity analysis on $\alphamax$, $\beta$, on the ImageNet $\rightarrow$ Caltech dataset. In this analysis, we set the following values for the other hyperparameters: $\eta_1 = 0.92$, $\eta_2 = 5.47$, $\varepsilon = 5.59$.
In the experiment on $\alphamax$, we set $\beta = 0.72$, while in the experiment on $\beta$, we set $\alphamax = 0.08$.

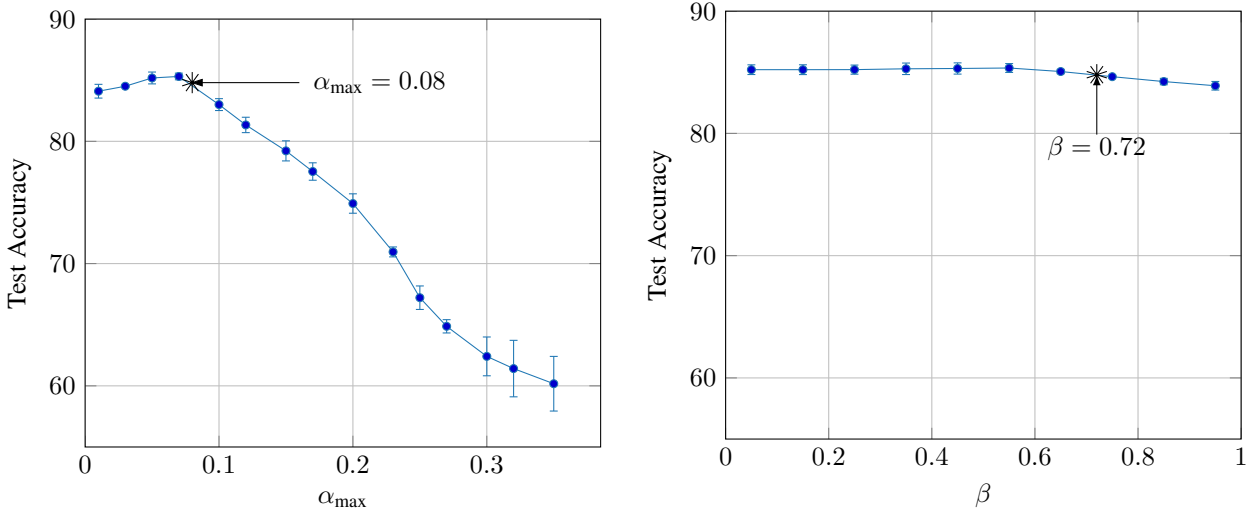
\begin{figure*}[t]
    \centering
    \begin{tabular}{ccc}
        \input{alpha_plot.tex} &
        \input{beta_plot.tex} &
    \end{tabular}
    \caption{The effect of changing alignment parameters $\alphamax$ and $\beta$ on test accuracy of ImageNet $\rightarrow$ Caltech.}
    \label{fig:sensitivity_imagenet}
\end{figure*}

As seen in \cref{fig:sensitivity_imagenet} (right), varying $\beta$ over the entire range $(0,1]$ has a limited impact on performance, with \update{variations} not exceeding 2\%. We observe a similar trend whenever $\alphamax$ is varied within the range $(0,0.1]$. When $\alphamax>0.1$, however, we see a significant drop in performance, caused most likely by the large number of outliers in the source sample. These results indicate that the specific choice of these parameters has a minor impact over a range of reasonable values.

\section{Relation between \algoname and ARPM Weights}\label{app:ARPM-weights}
The weights $\bp = (\bpedge{1}, \ldots, \bpedge{n_s})$ used in the ARPM algorithm of \citet{Gu2024} are defined as the solution of the following Wasserstein-1 type problem between the source and target distributions:
\begin{equation}
	\label{APP_objective}
	\min_{\bp\in \Delta}\mathbb{W}_1\!\left( \sum_{i=1}^{n_s} \tildepi \delta_{f(x_i)}, \dfrac{1}{n_t}\sum_{j=1}^{n_t} \delta_{f(\tilde{x}_j)} \right).
\end{equation}
Here, the constraint set $\Delta$ is defined as
\begin{subequations}
	\begin{empheq}[left={\Delta=\empheqlbrace}]{alignat = 2}
		& \tildepi \ge 0 \\
		& \sum_{i=1}^{n_s} \tildepi = 1 \\
		& \sum_{i=1}^{n_s} \left(\tildepi - \frac{1}{n_s}\right)^2 \leq \frac{\rho}{n_s} \label{square_ARPM}
	\end{empheq}
\end{subequations}
where $\rho$ is a hyperparameter.
Interestingly, the weights $\bp = (\bpedge{1}, \ldots, \bpedge{n_s})$ of \algoname, given in \eqref{eq:pi_tilde}, can be equivalently expressed, for the case $\alpha=1$, as the solution of the same optimization problem as in~\eqref{APP_objective}, but with the constraint set $\Delta$ replaced by $\Gamma$ \update{defined as:}
\begin{subequations}
	\begin{empheq}[left={\Gamma=\empheqlbrace}]{alignat = 2}
		& \tildepi \ge 0 \\
		& \sum_{i=1}^{n_s} \tildepi = 1 \\
		& \tildepi \leq \frac{1}{\beta n_s}. \label{Partial_mass}
	\end{empheq}
\end{subequations}
Here, $\beta$ is a hyperparameter.
Note that the only difference between the two sets of constraints is that they use different ways to control the magnitude of $\tildepi$ in \eqref{square_ARPM} and \eqref{Partial_mass}.
Compared to the ARPM constraint $\Delta$,
our constraint $\Gamma$ is not only simpler and more theoretically grounded,
but also more intuitive:
\eqref{Partial_mass} controls the maximum {target sample mass} that can be matched to a single source sample,
while the corresponding ARPM constraint \eqref{square_ARPM} is harder to interpret.
As demonstrated in our numerical results in \cref{table:office_main},
our weights also lead to better performance.

Note finally that ARPM solves~\eqref{APP_objective} only approximately:
it first solves a Wasserstein-1 problem with fixed $\tildepi = 1/n_s$ using a Wasserstein-GAN,
and then fixes this learned Wasserstein-GAN and updates the weights $\tildepi$.

\end{document}

%% file: tablesandfigures.tex
\newcommand{\tableweights}{
	\begin{table*}[t]
		\caption{Test accuracy on the Office-Home dataset using the weight choices described in \cref{sec:ablation_study}.}
		\vskip 0.1in
		\label{table:ablation_office}
		\centering
		\scalebox{0.63}{
			\begin{tabular}{lcccccccccccc|c}
				\toprule
				Weighting scheme & A$\rightarrow$C & A$\rightarrow$P & A$\rightarrow$R & C$\rightarrow$A & C$\rightarrow$P & C$\rightarrow$R & P$\rightarrow$A & P$\rightarrow$C & P$\rightarrow$R & R$\rightarrow$A & R$\rightarrow$C & R$\rightarrow$P & Avg           \\
				\midrule
				MPOT weights     & 62.2 (1.4) & 81.2 (1.0) & 88.6 (0.6) & 73.0 (1.1) & 77.0 (3.1) & 79.2 (1.3) & 74.4 (1.3) & 61.7 (0.9) & 85.1 (1.5) & 80.0 (0.7) & 65.9 (0.5) & 83.8 (0.5) & 76.0 (0.4) \\
				BA$^3$US weights & \textbf{64.2} (1.5) & 82.6 (2.0) & 89.2 (0.4) & \textbf{75.5} (0.8) & 76.4 (4.1) & 80.2 (0.9) & \textbf{76.6} (1.4) & \textbf{62.9} (1.4) & \textbf{88.5} (1.1) & \textbf{81.3} (0.8) & \textbf{67.4} (0.6) & \textbf{86.7} (0.9) & \textbf{77.6} (0.4) \\
				ARPM weights     & 55.9 (1.0) & 75.8 (0.6) & 87.1 (0.4) & 69.7 (1.5) & 73.7 (1.4) & 75.2 (1.1) & 72.4 (1.6) & 55.8 (2.6) & 81.2 (1.0) & 80.5 (0.6) & 63.7 (0.8) & 83.9 (0.7) & 72.9 (0.3) \\
				\midrule
				\algoname (ours) & 62.5 (1.2) & \textbf{83.0} (1.1) & \textbf{89.5} (0.3) & 75.2 (1.1) & \textbf{78.4} (2.2) & \textbf{82.3} (1.3) & \textbf{76.6} (1.5) & 61.4 (3.1) & 88.0 (1.0) & 81.1 (0.7) & 66.5 (1.0) & 86.6 (0.6) & \textbf{77.6} (0.7) \\
				\bottomrule
			\end{tabular}
		}
	\end{table*}
}

\newcommand{\tableweightsimagenet}{
	\begin{table*}[t]
		\caption{Test accuracy on the ImageNet $\rightarrow$ Caltech dataset using the weight choices described in \cref{sec:ablation_study}.}
		\vskip 0.1in
		\label{table:ablation_imagenet}
		\centering
		\scalebox{0.79}{
			\begin{tabular}{l|c}
				\toprule
				Weighting scheme       & Test accuracy \\
				\midrule
				MPOT weights           & 78.6 (1.2) \\
				BA$^3$US weights & 84.7 (0.7) \\
				ARPM weights     & 79.2 (1.4) \\
				\midrule
				\algoname (ours) & \textbf{84.8} (0.1) \\
				\bottomrule
			\end{tabular}
		}
	\end{table*}
}

\newcommand{\tableImageNet}{
	\begin{table*}[t]
		\caption{Test accuracy on the ImageNet $\rightarrow$ Caltech dataset.}
		\vskip 0.1in
		\label{table:imagenet_main}
		\centering
		\scalebox{0.79}{
			\begin{tabular}{l|c}
				\toprule
				Algorithm                             & Test accuracy \\
				\midrule
				ResNet-50~\cite{He_2016_CVPR}         & 69.7             \\
				DAN~\cite{long2015learning}           & 71.3             \\
				DANN~\cite{ganin2016domain}           & 70.8             \\
				IWAN~\cite{zhang2018importance}       & 78.1             \\
				PADA~\cite{cao2018partial}            & 75.0             \\
				ETN~\cite{cao2019learning}            & 83.2             \\
				DRCN~\cite{li2020deep}                & 75.3             \\
				BA$^3$US~\cite{liang2020balanced}     & 84.0             \\
				ISRA+BA$^3$US~\cite{xiao2021implicit} & 85.3             \\
				SLM~\cite{sahoo2021select}            & 82.3             \\
				SAN++~\cite{9736609}                  & 83.3             \\
				AR~\cite{gu2021adversarial}           & 85.4 (0.2)       \\
				ARPM \citep{Gu2024}                   & 84.1 (1.4)       \\
				\midrule
				PWAN \citep{wang2024partial}          & \textbf{86.0 (0.5)}      \\
				\midrule
				\algoname (ours)                      & 84.8 (0.1)      \\
				\algonamearpm                         & 85.1 (0.9)      \\
				\bottomrule
			\end{tabular}
		}
	\end{table*}
}

\newcommand{\tableOfficeHome}{
	\begin{table*}[t]
		\caption{Test accuracy on the Office-Home dataset.}
		\vskip 0.1in
		\label{table:office_main}
		\centering
		\scalebox{0.58}{
			\begin{tabular}{lcccccccccccc|c}
				\toprule
				Algorithm                             & A$\rightarrow$C & A$\rightarrow$P & A$\rightarrow$R & C$\rightarrow$A & C$\rightarrow$P & C$\rightarrow$R & P$\rightarrow$A & P$\rightarrow$C & P$\rightarrow$R & R$\rightarrow$A & R$\rightarrow$C & R$\rightarrow$P & Avg           \\
				\midrule
				ResNet-50~\cite{He_2016_CVPR}         & 46.3            & 67.5            & 75.9            & 59.1            & 59.9            & 62.7            & 58.2            & 41.8            & 74.9            & 67.4            & 48.2            & 74.2            & 61.4          \\
				ADDA~\cite{tzeng2017adversarial}      & 45.2            & 68.8            & 79.2            & 64.6            & 60.0            & 68.3            & 57.6            & 38.9            & 77.5            & 70.3            & 45.2            & 78.3            & 62.8          \\
				CDAN+E~\cite{NIPS2018_7436}           & 47.5            & 65.9            & 75.7            & 57.1            & 54.1            & 63.4            & 59.6            & 44.3            & 72.4            & 66.0            & 49.9            & 72.8            & 60.7          \\
				IWAN~\cite{zhang2018importance}       & 53.9            & 54.5            & 78.1            & 61.3            & 48.0            & 63.3            & 54.2            & 52.0            & 81.3            & 76.5            & 56.8            & 82.9            & 63.6          \\
				PADA~\cite{cao2018partial}            & 52.0            & 67.0            & 78.7            & 52.2            & 53.8            & 59.0            & 52.6            & 43.2            & 78.8            & 73.7            & 56.6            & 77.1            & 62.1          \\
				ETN~\cite{cao2019learning}            & 59.2            & 77.0            & 79.5            & 62.9            & 65.7            & 75.0            & 68.3            & 55.4            & 84.4            & 75.7            & 57.7            & 84.5            & 70.5          \\
				DRCN~\cite{li2020deep}                & 54.0            & 76.4            & 83.0            & 62.1            & 64.5            & 71.0            & 70.8            & 49.8            & 80.5            & 77.5            & 59.1            & 79.9            & 69.0          \\
				BA$^3$US~\cite{liang2020balanced}     & 60.6            & 83.2            & 88.4            & 71.8            & 72.8            & 83.4            & 75.5            & 61.6            & 86.5            & 79.3            & 62.8            & 86.1            & 76.0          \\
				ISRA+BA$^3$US~\cite{xiao2021implicit} & {64.7}          & 83.0            & 89.1            & 75.7            & 75.5            & 85.4            & 78.5            & {64.2}          & 88.1            & 81.3            & 65.3            & 86.7            & 78.2          \\
				SHOT++~\cite{9512429}                 & {65.0}          & 85.8            & \bf93.4         & 78.8            & 77.4            & {87.3}          & 79.3            & {66.0}          & 89.6            & 81.3            & {68.1}          & 86.8            & 79.9          \\
				SPDA~\cite{guo2022selective}          & 64.2            & 87.8        & 88.0            & 74.3            & 75.1            & 79.1            & 79.4            & 58.9            & 85.1            & 81.4            & 67.4            & 84.1            & 77.1          \\
				APDA-CI~\cite{lin2022adversarial}     & 61.7            & {86.9}          & 90.5            & 77.2            & 76.9            & 83.8            & 79.6            & 63.8            & 88.5            & {85.0}          & 65.8            & 86.2            & 78.8          \\
				CLA~\cite{9705553}                    & 66.7            & 85.6            & 90.9            & 75.6            & 76.9            & 86.8            & 78.8            & {67.4}          & 88.7            & 81.7            & 66.9            & 87.8            & 79.5          \\
				RAN~\cite{9957101}                    & 63.3            & 83.1            & 89.0            & 75.0            & 74.5            & 82.9            & 78.0            & 61.2            & 86.7            & 79.9            & 63.5            & 85.0            & 76.8          \\
				JUMBOT~\cite{fatras2021unbalanced}                 & 62.7            & 77.5            & 84.4            & 76.0            & 73.3            & 80.5            & 74.7            & 60.8            & 85.1            & 80.2            & 66.5            & 83.9            & 75.5          \\
				STCPDA~\cite{he2023addressing}        & 63.1            & 87.8         & 90.1            & 77.2            & 75.4            & 85.6            & \bf 81.4        & 62.4            & \bf 90.5        & 82.6            & {69.5}          & 88.2            & 79.5          \\
				SLM~\cite{sahoo2021select}            & 61.1            & 84.0            & 91.4            & 76.5            & 75.0            & 81.8            & 74.6            & 55.6            & 87.8            & 82.3            & 57.8            & 83.5            & 76.0          \\
				SAN++~\cite{9736609}                  & 61.3            & 81.6            & 88.6            & 72.8            & 76.4            & 81.9            & 74.5            & 57.7            & 87.2            & 79.7            & 63.8            & 86.1            & 76.0          \\
				IDSP~\cite{li2022partial}             & 60.8            & 80.8            & 87.3            & 69.3            & 76.0            & 80.2            & 74.7            & 59.2            & 85.3            & 77.8            & 61.3            & 85.7            & 74.9          \\
				MOT~\cite{Luo_2023_CVPR}              & 63.1            & {86.1}          & {92.3}          & {78.7}          & \bf85.4         & \bf89.6         & {79.8}          & 62.3            & {89.7}          & {83.8}          & {67.0}          & 89.6         & {80.6}        \\
				AR~\cite{gu2021adversarial}           & {67.4}          & 85.3            & 90.0            & 77.3            & 70.6            & 85.2            & 79.0            & 64.8            & 89.5            & 80.4            & 66.2            & 86.4            & 78.3          \\
				ARPM \citep{Gu2024}                   & 67.4 (2.5)      & \textbf{88.4} (1.4) & 92.7 (1.1) & 79.9 (2.1) & 82.6 (2.6) & 87.0 (0.7) & 78.8 (3.1) & \textbf{69.1} (1.6) & 89.6 (0.3) & 86.0 (1.0) & 69.5 (2.2) & \textbf{89.7} (1.1) & 81.7 (0.7)          \\
				\midrule
				MPOT \citep{nguyen2022improving}      & 64.6            & 80.6            & 87.1            & 76.4            & 77.6            & 83.5            & 77.0   & 63.7            & 87.6            & 81.4   & 68.5   & 87.3   & 77.9          \\
				PWAN \citep{wang2024partial}          & 63.3 (1.8) & 84.1 (1.8) & 89.3 (0.6) & 76.7 (1.1) & 75.6 (2.0) & 83.8 (1.8) & 76.6 (0.8) & 60.7 (2.2) & 86.7 (0.8) & 80.1 (0.6) & 64.4 (0.5) & 86.6 (0.7) & 77.3 (0.5)          \\
				\midrule
                \algoname (ours)                      & 62.5 (1.2)      & 83.0 (1.1)      & 89.5 (0.3)      & 75.2 (1.1)      & 78.4 (2.2)      & 82.3 (1.3)      & 76.6 (1.5)      & 61.4 (3.1)      & 88.0 (1.0)      & 81.1 (0.7)      & 66.5 (1.0)      & 86.6 (0.6)      & 77.6 (0.7)    \\
				\algonamearpm                         & \textbf{69.0} (1.9) & 87.2 (1.7)  & 92.8 (0.7)      & \textbf{81.0} (1.1) & 83.4 (2.7) & 86.0 (2.8) & 79.9 (2.2) & \textbf{69.1} (0.8) & 90.2 (0.9) & \textbf{86.6} (0.8) & \textbf{69.7} (2.3) & 88.7 (1.0) & \textbf{82.0} (0.4) \\
				\bottomrule
			\end{tabular}
		}
	\end{table*}
}

%% file: weights_plot.tex
\begin{tikzpicture}

\definecolor{crimson2143940}{RGB}{214,39,40}
\definecolor{darkgray176}{RGB}{176,176,176}
\definecolor{forestgreen4416044}{RGB}{44,160,44}
\definecolor{steelblue31119180}{RGB}{31,119,180}

\begin{groupplot}[group style={group size=3 by 1, horizontal sep=1.5cm}]
\nextgroupplot[
    title={All weights},
    tick align=outside,
    tick pos=left,
    title style={font=\large},
    xmin=0.0, xmax=1.0,
    ymin=0, ymax=1000,
]
\draw[draw=black,fill=steelblue31119180] (axis cs:0,0) rectangle (axis cs:0.02,2116);
\draw[draw=black,fill=steelblue31119180] (axis cs:0.0199999995529652,0) rectangle (axis cs:0.0399999995529652,233);
\draw[draw=black,fill=steelblue31119180] (axis cs:0.0399999991059303,0) rectangle (axis cs:0.0599999991059303,117);
\draw[draw=black,fill=steelblue31119180] (axis cs:0.0599999986588955,0) rectangle (axis cs:0.0799999986588955,33);
\draw[draw=black,fill=steelblue31119180] (axis cs:0.0799999982118607,0) rectangle (axis cs:0.0999999982118607,106);
\draw[draw=black,fill=steelblue31119180] (axis cs:0.0999999940395355,0) rectangle (axis cs:0.119999994039536,86);
\draw[draw=black,fill=steelblue31119180] (axis cs:0.119999997317791,0) rectangle (axis cs:0.139999997317791,60);
\draw[draw=black,fill=steelblue31119180] (axis cs:0.140000000596046,0) rectangle (axis cs:0.160000000596046,26);
\draw[draw=black,fill=steelblue31119180] (axis cs:0.159999996423721,0) rectangle (axis cs:0.179999996423721,0);
\draw[draw=black,fill=steelblue31119180] (axis cs:0.179999992251396,0) rectangle (axis cs:0.199999992251396,0);
\draw[draw=black,fill=steelblue31119180] (axis cs:0.199999988079071,0) rectangle (axis cs:0.219999988079071,0);
\draw[draw=black,fill=steelblue31119180] (axis cs:0.219999998807907,0) rectangle (axis cs:0.239999998807907,0);
\draw[draw=black,fill=steelblue31119180] (axis cs:0.239999994635582,0) rectangle (axis cs:0.259999994635582,0);
\draw[draw=black,fill=steelblue31119180] (axis cs:0.259999990463257,0) rectangle (axis cs:0.279999990463257,0);
\draw[draw=black,fill=steelblue31119180] (axis cs:0.280000001192093,0) rectangle (axis cs:0.300000001192093,1);
\draw[draw=black,fill=steelblue31119180] (axis cs:0.299999982118607,0) rectangle (axis cs:0.319999982118607,5);
\draw[draw=black,fill=steelblue31119180] (axis cs:0.319999992847443,0) rectangle (axis cs:0.339999992847443,57);
\draw[draw=black,fill=steelblue31119180] (axis cs:0.340000003576279,0) rectangle (axis cs:0.360000003576279,34);
\draw[draw=black,fill=steelblue31119180] (axis cs:0.359999984502792,0) rectangle (axis cs:0.379999984502792,6);
\draw[draw=black,fill=steelblue31119180] (axis cs:0.379999995231628,0) rectangle (axis cs:0.399999995231628,53);
\draw[draw=black,fill=steelblue31119180] (axis cs:0.399999976158142,0) rectangle (axis cs:0.419999976158142,21);
\draw[draw=black,fill=steelblue31119180] (axis cs:0.419999986886978,0) rectangle (axis cs:0.439999986886978,49);
\draw[draw=black,fill=steelblue31119180] (axis cs:0.439999997615814,0) rectangle (axis cs:0.459999997615814,1);
\draw[draw=black,fill=steelblue31119180] (axis cs:0.459999978542328,0) rectangle (axis cs:0.479999978542328,5);
\draw[draw=black,fill=steelblue31119180] (axis cs:0.479999989271164,0) rectangle (axis cs:0.499999989271164,68);
\draw[draw=black,fill=steelblue31119180] (axis cs:0.5,0) rectangle (axis cs:0.52,92);
\draw[draw=black,fill=steelblue31119180] (axis cs:0.519999980926514,0) rectangle (axis cs:0.539999980926514,16);
\draw[draw=black,fill=steelblue31119180] (axis cs:0.540000021457672,0) rectangle (axis cs:0.560000021457672,41);
\draw[draw=black,fill=steelblue31119180] (axis cs:0.560000002384186,0) rectangle (axis cs:0.580000002384186,27);
\draw[draw=black,fill=steelblue31119180] (axis cs:0.579999983310699,0) rectangle (axis cs:0.599999983310699,46);
\draw[draw=black,fill=steelblue31119180] (axis cs:0.600000023841858,0) rectangle (axis cs:0.620000023841858,174);
\draw[draw=black,fill=steelblue31119180] (axis cs:0.620000004768372,0) rectangle (axis cs:0.640000004768372,41);
\draw[draw=black,fill=steelblue31119180] (axis cs:0.639999985694885,0) rectangle (axis cs:0.659999985694885,0);
\draw[draw=black,fill=steelblue31119180] (axis cs:0.660000026226044,0) rectangle (axis cs:0.680000026226044,6);
\draw[draw=black,fill=steelblue31119180] (axis cs:0.680000007152557,0) rectangle (axis cs:0.700000007152557,43);
\draw[draw=black,fill=steelblue31119180] (axis cs:0.699999988079071,0) rectangle (axis cs:0.719999988079071,86);
\draw[draw=black,fill=steelblue31119180] (axis cs:0.720000028610229,0) rectangle (axis cs:0.74000002861023,115);
\draw[draw=black,fill=steelblue31119180] (axis cs:0.740000009536743,0) rectangle (axis cs:0.760000009536743,61);
\draw[draw=black,fill=steelblue31119180] (axis cs:0.759999990463257,0) rectangle (axis cs:0.779999990463257,59);
\draw[draw=black,fill=steelblue31119180] (axis cs:0.780000030994415,0) rectangle (axis cs:0.800000030994415,46);
\draw[draw=black,fill=steelblue31119180] (axis cs:0.800000011920929,0) rectangle (axis cs:0.820000011920929,89);
\draw[draw=black,fill=steelblue31119180] (axis cs:0.819999992847443,0) rectangle (axis cs:0.839999992847443,45);
\draw[draw=black,fill=steelblue31119180] (axis cs:0.840000033378601,0) rectangle (axis cs:0.860000033378601,3);
\draw[draw=black,fill=steelblue31119180] (axis cs:0.860000014305115,0) rectangle (axis cs:0.880000014305115,36);
\draw[draw=black,fill=steelblue31119180] (axis cs:0.879999995231628,0) rectangle (axis cs:0.899999995231628,4);
\draw[draw=black,fill=steelblue31119180] (axis cs:0.899999976158142,0) rectangle (axis cs:0.919999976158142,1);
\draw[draw=black,fill=steelblue31119180] (axis cs:0.920000016689301,0) rectangle (axis cs:0.940000016689301,3);
\draw[draw=black,fill=steelblue31119180] (axis cs:0.939999997615814,0) rectangle (axis cs:0.959999997615814,4);
\draw[draw=black,fill=steelblue31119180] (axis cs:0.959999978542328,0) rectangle (axis cs:0.979999978542328,4);
\draw[draw=black,fill=steelblue31119180] (axis cs:0.980000019073486,0) rectangle (axis cs:1.00000001907349,320);

\nextgroupplot[
    tick align=outside,
    tick pos=left,
    align=center,
    title style={font=\large},
    title={{Shared classes\\Percentage of total weights = 93.96\%}},
    xmin=0.0, xmax=1.0,
    ymin=0, ymax=1000,
]
\draw[draw=black,fill=forestgreen4416044] (axis cs:0,0) rectangle (axis cs:0.02,26);
\draw[draw=black,fill=forestgreen4416044] (axis cs:0.0199999995529652,0) rectangle (axis cs:0.0399999995529652,0);
\draw[draw=black,fill=forestgreen4416044] (axis cs:0.0399999991059303,0) rectangle (axis cs:0.0599999991059303,0);
\draw[draw=black,fill=forestgreen4416044] (axis cs:0.0599999986588955,0) rectangle (axis cs:0.0799999986588955,32);
\draw[draw=black,fill=forestgreen4416044] (axis cs:0.0799999982118607,0) rectangle (axis cs:0.0999999982118607,9);
\draw[draw=black,fill=forestgreen4416044] (axis cs:0.0999999940395355,0) rectangle (axis cs:0.119999994039536,39);
\draw[draw=black,fill=forestgreen4416044] (axis cs:0.119999997317791,0) rectangle (axis cs:0.139999997317791,60);
\draw[draw=black,fill=forestgreen4416044] (axis cs:0.140000000596046,0) rectangle (axis cs:0.160000000596046,26);
\draw[draw=black,fill=forestgreen4416044] (axis cs:0.159999996423721,0) rectangle (axis cs:0.179999996423721,0);
\draw[draw=black,fill=forestgreen4416044] (axis cs:0.179999992251396,0) rectangle (axis cs:0.199999992251396,0);
\draw[draw=black,fill=forestgreen4416044] (axis cs:0.199999988079071,0) rectangle (axis cs:0.219999988079071,0);
\draw[draw=black,fill=forestgreen4416044] (axis cs:0.219999998807907,0) rectangle (axis cs:0.239999998807907,0);
\draw[draw=black,fill=forestgreen4416044] (axis cs:0.239999994635582,0) rectangle (axis cs:0.259999994635582,0);
\draw[draw=black,fill=forestgreen4416044] (axis cs:0.259999990463257,0) rectangle (axis cs:0.279999990463257,0);
\draw[draw=black,fill=forestgreen4416044] (axis cs:0.280000001192093,0) rectangle (axis cs:0.300000001192093,1);
\draw[draw=black,fill=forestgreen4416044] (axis cs:0.299999982118607,0) rectangle (axis cs:0.319999982118607,5);
\draw[draw=black,fill=forestgreen4416044] (axis cs:0.319999992847443,0) rectangle (axis cs:0.339999992847443,57);
\draw[draw=black,fill=forestgreen4416044] (axis cs:0.340000003576279,0) rectangle (axis cs:0.360000003576279,34);
\draw[draw=black,fill=forestgreen4416044] (axis cs:0.359999984502792,0) rectangle (axis cs:0.379999984502792,6);
\draw[draw=black,fill=forestgreen4416044] (axis cs:0.379999995231628,0) rectangle (axis cs:0.399999995231628,53);
\draw[draw=black,fill=forestgreen4416044] (axis cs:0.399999976158142,0) rectangle (axis cs:0.419999976158142,21);
\draw[draw=black,fill=forestgreen4416044] (axis cs:0.419999986886978,0) rectangle (axis cs:0.439999986886978,49);
\draw[draw=black,fill=forestgreen4416044] (axis cs:0.439999997615814,0) rectangle (axis cs:0.459999997615814,1);
\draw[draw=black,fill=forestgreen4416044] (axis cs:0.459999978542328,0) rectangle (axis cs:0.479999978542328,5);
\draw[draw=black,fill=forestgreen4416044] (axis cs:0.479999989271164,0) rectangle (axis cs:0.499999989271164,68);
\draw[draw=black,fill=forestgreen4416044] (axis cs:0.5,0) rectangle (axis cs:0.52,92);
\draw[draw=black,fill=forestgreen4416044] (axis cs:0.519999980926514,0) rectangle (axis cs:0.539999980926514,16);
\draw[draw=black,fill=forestgreen4416044] (axis cs:0.540000021457672,0) rectangle (axis cs:0.560000021457672,41);
\draw[draw=black,fill=forestgreen4416044] (axis cs:0.560000002384186,0) rectangle (axis cs:0.580000002384186,19);
\draw[draw=black,fill=forestgreen4416044] (axis cs:0.579999983310699,0) rectangle (axis cs:0.599999983310699,20);
\draw[draw=black,fill=forestgreen4416044] (axis cs:0.600000023841858,0) rectangle (axis cs:0.620000023841858,165);
\draw[draw=black,fill=forestgreen4416044] (axis cs:0.620000004768372,0) rectangle (axis cs:0.640000004768372,41);
\draw[draw=black,fill=forestgreen4416044] (axis cs:0.639999985694885,0) rectangle (axis cs:0.659999985694885,0);
\draw[draw=black,fill=forestgreen4416044] (axis cs:0.660000026226044,0) rectangle (axis cs:0.680000026226044,6);
\draw[draw=black,fill=forestgreen4416044] (axis cs:0.680000007152557,0) rectangle (axis cs:0.700000007152557,43);
\draw[draw=black,fill=forestgreen4416044] (axis cs:0.699999988079071,0) rectangle (axis cs:0.719999988079071,86);
\draw[draw=black,fill=forestgreen4416044] (axis cs:0.720000028610229,0) rectangle (axis cs:0.74000002861023,115);
\draw[draw=black,fill=forestgreen4416044] (axis cs:0.740000009536743,0) rectangle (axis cs:0.760000009536743,61);
\draw[draw=black,fill=forestgreen4416044] (axis cs:0.759999990463257,0) rectangle (axis cs:0.779999990463257,59);
\draw[draw=black,fill=forestgreen4416044] (axis cs:0.780000030994415,0) rectangle (axis cs:0.800000030994415,46);
\draw[draw=black,fill=forestgreen4416044] (axis cs:0.800000011920929,0) rectangle (axis cs:0.820000011920929,89);
\draw[draw=black,fill=forestgreen4416044] (axis cs:0.819999992847443,0) rectangle (axis cs:0.839999992847443,45);
\draw[draw=black,fill=forestgreen4416044] (axis cs:0.840000033378601,0) rectangle (axis cs:0.860000033378601,3);
\draw[draw=black,fill=forestgreen4416044] (axis cs:0.860000014305115,0) rectangle (axis cs:0.880000014305115,36);
\draw[draw=black,fill=forestgreen4416044] (axis cs:0.879999995231628,0) rectangle (axis cs:0.899999995231628,4);
\draw[draw=black,fill=forestgreen4416044] (axis cs:0.899999976158142,0) rectangle (axis cs:0.919999976158142,1);
\draw[draw=black,fill=forestgreen4416044] (axis cs:0.920000016689301,0) rectangle (axis cs:0.940000016689301,3);
\draw[draw=black,fill=forestgreen4416044] (axis cs:0.939999997615814,0) rectangle (axis cs:0.959999997615814,4);
\draw[draw=black,fill=forestgreen4416044] (axis cs:0.959999978542328,0) rectangle (axis cs:0.979999978542328,4);
\draw[draw=black,fill=forestgreen4416044] (axis cs:0.980000019073486,0) rectangle (axis cs:1.00000001907349,320);

\nextgroupplot[
    tick align=outside,
    tick pos=left,
    align=center,
    title style={font=\large},
    title={{Outlier classes\\Percentage of total weights = 6.04\%}},
    xmin=0.0, xmax=1.0,
    ymin=0, ymax=1000,
]
\draw[draw=black,fill=crimson2143940] (axis cs:0,0) rectangle (axis cs:0.02,2090);
\draw[draw=black,fill=crimson2143940] (axis cs:0.0199999995529652,0) rectangle (axis cs:0.0399999995529652,233);
\draw[draw=black,fill=crimson2143940] (axis cs:0.0399999991059303,0) rectangle (axis cs:0.0599999991059303,117);
\draw[draw=black,fill=crimson2143940] (axis cs:0.0599999986588955,0) rectangle (axis cs:0.0799999986588955,1);
\draw[draw=black,fill=crimson2143940] (axis cs:0.0799999982118607,0) rectangle (axis cs:0.0999999982118607,97);
\draw[draw=black,fill=crimson2143940] (axis cs:0.0999999940395355,0) rectangle (axis cs:0.119999994039536,47);
\draw[draw=black,fill=crimson2143940] (axis cs:0.119999997317791,0) rectangle (axis cs:0.139999997317791,0);
\draw[draw=black,fill=crimson2143940] (axis cs:0.140000000596046,0) rectangle (axis cs:0.160000000596046,0);
\draw[draw=black,fill=crimson2143940] (axis cs:0.159999996423721,0) rectangle (axis cs:0.179999996423721,0);
\draw[draw=black,fill=crimson2143940] (axis cs:0.179999992251396,0) rectangle (axis cs:0.199999992251396,0);
\draw[draw=black,fill=crimson2143940] (axis cs:0.199999988079071,0) rectangle (axis cs:0.219999988079071,0);
\draw[draw=black,fill=crimson2143940] (axis cs:0.219999998807907,0) rectangle (axis cs:0.239999998807907,0);
\draw[draw=black,fill=crimson2143940] (axis cs:0.239999994635582,0) rectangle (axis cs:0.259999994635582,0);
\draw[draw=black,fill=crimson2143940] (axis cs:0.259999990463257,0) rectangle (axis cs:0.279999990463257,0);
\draw[draw=black,fill=crimson2143940] (axis cs:0.280000001192093,0) rectangle (axis cs:0.300000001192093,0);
\draw[draw=black,fill=crimson2143940] (axis cs:0.299999982118607,0) rectangle (axis cs:0.319999982118607,0);
\draw[draw=black,fill=crimson2143940] (axis cs:0.319999992847443,0) rectangle (axis cs:0.339999992847443,0);
\draw[draw=black,fill=crimson2143940] (axis cs:0.340000003576279,0) rectangle (axis cs:0.360000003576279,0);
\draw[draw=black,fill=crimson2143940] (axis cs:0.359999984502792,0) rectangle (axis cs:0.379999984502792,0);
\draw[draw=black,fill=crimson2143940] (axis cs:0.379999995231628,0) rectangle (axis cs:0.399999995231628,0);
\draw[draw=black,fill=crimson2143940] (axis cs:0.399999976158142,0) rectangle (axis cs:0.419999976158142,0);
\draw[draw=black,fill=crimson2143940] (axis cs:0.419999986886978,0) rectangle (axis cs:0.439999986886978,0);
\draw[draw=black,fill=crimson2143940] (axis cs:0.439999997615814,0) rectangle (axis cs:0.459999997615814,0);
\draw[draw=black,fill=crimson2143940] (axis cs:0.459999978542328,0) rectangle (axis cs:0.479999978542328,0);
\draw[draw=black,fill=crimson2143940] (axis cs:0.479999989271164,0) rectangle (axis cs:0.499999989271164,0);
\draw[draw=black,fill=crimson2143940] (axis cs:0.5,0) rectangle (axis cs:0.52,0);
\draw[draw=black,fill=crimson2143940] (axis cs:0.519999980926514,0) rectangle (axis cs:0.539999980926514,0);
\draw[draw=black,fill=crimson2143940] (axis cs:0.540000021457672,0) rectangle (axis cs:0.560000021457672,0);
\draw[draw=black,fill=crimson2143940] (axis cs:0.560000002384186,0) rectangle (axis cs:0.580000002384186,8);
\draw[draw=black,fill=crimson2143940] (axis cs:0.579999983310699,0) rectangle (axis cs:0.599999983310699,26);
\draw[draw=black,fill=crimson2143940] (axis cs:0.600000023841858,0) rectangle (axis cs:0.620000023841858,9);
\draw[draw=black,fill=crimson2143940] (axis cs:0.620000004768372,0) rectangle (axis cs:0.640000004768372,0);
\draw[draw=black,fill=crimson2143940] (axis cs:0.639999985694885,0) rectangle (axis cs:0.659999985694885,0);
\draw[draw=black,fill=crimson2143940] (axis cs:0.660000026226044,0) rectangle (axis cs:0.680000026226044,0);
\draw[draw=black,fill=crimson2143940] (axis cs:0.680000007152557,0) rectangle (axis cs:0.700000007152557,0);
\draw[draw=black,fill=crimson2143940] (axis cs:0.699999988079071,0) rectangle (axis cs:0.719999988079071,0);
\draw[draw=black,fill=crimson2143940] (axis cs:0.720000028610229,0) rectangle (axis cs:0.74000002861023,0);
\draw[draw=black,fill=crimson2143940] (axis cs:0.740000009536743,0) rectangle (axis cs:0.760000009536743,0);
\draw[draw=black,fill=crimson2143940] (axis cs:0.759999990463257,0) rectangle (axis cs:0.779999990463257,0);
\draw[draw=black,fill=crimson2143940] (axis cs:0.780000030994415,0) rectangle (axis cs:0.800000030994415,0);
\draw[draw=black,fill=crimson2143940] (axis cs:0.800000011920929,0) rectangle (axis cs:0.820000011920929,0);
\draw[draw=black,fill=crimson2143940] (axis cs:0.819999992847443,0) rectangle (axis cs:0.839999992847443,0);
\draw[draw=black,fill=crimson2143940] (axis cs:0.840000033378601,0) rectangle (axis cs:0.860000033378601,0);
\draw[draw=black,fill=crimson2143940] (axis cs:0.860000014305115,0) rectangle (axis cs:0.880000014305115,0);
\draw[draw=black,fill=crimson2143940] (axis cs:0.879999995231628,0) rectangle (axis cs:0.899999995231628,0);
\draw[draw=black,fill=crimson2143940] (axis cs:0.899999976158142,0) rectangle (axis cs:0.919999976158142,0);
\draw[draw=black,fill=crimson2143940] (axis cs:0.920000016689301,0) rectangle (axis cs:0.940000016689301,0);
\draw[draw=black,fill=crimson2143940] (axis cs:0.939999997615814,0) rectangle (axis cs:0.959999997615814,0);
\draw[draw=black,fill=crimson2143940] (axis cs:0.959999978542328,0) rectangle (axis cs:0.979999978542328,0);
\draw[draw=black,fill=crimson2143940] (axis cs:0.980000019073486,0) rectangle (axis cs:1.00000001907349,0);
\end{groupplot}

\end{tikzpicture}

%% file: alpha_plot.tex
\begin{tikzpicture}

\definecolor{steelblue31119180}{RGB}{31,119,180}

\pgfplotstableread[col sep=comma]{alpha_data.csv}\alphatable

\begin{axis}[
    xlabel={$\alphamax$},
    ylabel={Test Accuracy},
    ymin=55,
    ymax=90,
    grid=major,
    xmin=0,
]
    \addplot+ [
        steelblue31119180,
        mark size=1.5pt,
        error bars, y dir=both, y explicit,
    ] table [x=alpha_max,y=mean,y error=std] {\alphatable};
    \addplot+ [
        black,
        mark=10-pointed star,
        mark size=4pt,
    ] coordinates {(0.08, 84.8)};
    \node at (axis cs: 0.22, 84.8) {$\alphamax = 0.08$};
    \draw[-latex] (axis cs: 0.16, 84.8)--(axis cs: 0.08, 84.8);
\end{axis}

\end{tikzpicture}

%% file: beta_plot.tex
\begin{tikzpicture}

\definecolor{steelblue31119180}{RGB}{31,119,180}

\pgfplotstableread[col sep=comma]{beta_data.csv}\betatable

\begin{axis}[
    xlabel={$\beta$},
    ylabel={Test Accuracy},
    ymin=55,
    ymax=90,
    grid=major,
    xmin=0,
    xmax=1,
]
    \addplot+ [
        steelblue31119180,
        mark size=1.5pt,
        error bars, y dir=both, y explicit,
    ] table [x=beta,y=mean,y error=std] {\betatable};
    \addplot+ [
        black,
        mark=10-pointed star,
        mark size=4pt,
    ] coordinates {(0.72, 84.8)};
    \node at (axis cs: 0.72, 78.8) {$\beta = 0.72$};
    \draw[-latex] (axis cs: 0.72, 79.9)--(axis cs: 0.72, 84.8);
\end{axis}

\end{tikzpicture}

%% file: main.bbl
\begin{thebibliography}{58}
\providecommand{\natexlab}[1]{#1}
\providecommand{\url}[1]{\texttt{#1}}
\expandafter\ifx\csname urlstyle\endcsname\relax
  \providecommand{\doi}[1]{doi: #1}\else
  \providecommand{\doi}{doi: \begingroup \urlstyle{rm}\Url}\fi

\bibitem[Alquier(2024)]{alquier2021user}
Alquier, P.
\newblock User-friendly introduction to {PAC-Bayes} bounds.
\newblock \emph{Found. Trends Mach. Learn.}, 17\penalty0 (2):\penalty0 174--303, Jan. 2024.

\bibitem[Arjovsky et~al.(2017)Arjovsky, Chintala, and Bottou]{arjovsky2017wasserstein}
Arjovsky, M., Chintala, S., and Bottou, L.
\newblock Wasserstein generative adversarial networks.
\newblock In \emph{Proc. Int. Conf. Mach. Learning (ICML)}, Sydney, Australia, Aug. 2017.

\bibitem[Ben-David et~al.(2006)Ben-David, Blitzer, Crammer, and Pereira]{ben2006analysis}
Ben-David, S., Blitzer, J., Crammer, K., and Pereira, F.
\newblock Analysis of representations for domain adaptation.
\newblock In \emph{Proc. Conf. Neural Inf. Process. Syst. (NeurIPS)}, Vancouver, Canada, Dec. 2006.

\bibitem[Ben-David et~al.(2010)Ben-David, Blitzer, Crammer, Kulesza, Pereira, and Vaughan]{ben2010theory}
Ben-David, S., Blitzer, J., Crammer, K., Kulesza, A., Pereira, F., and Vaughan, J.~W.
\newblock A theory of learning from different domains.
\newblock \emph{Mach. Learn.}, 79\penalty0 (1):\penalty0 151--175, Oct. 2010.

\bibitem[Boyd \& Vandenberghe(2004)Boyd and Vandenberghe]{boyd04-a}
Boyd, S. and Vandenberghe, L.
\newblock \emph{Convex Optimization}.
\newblock Cambridge Univ. Press, Cambridge, UK, 2004.

\bibitem[Caffarelli \& McCann(2010)Caffarelli and McCann]{caffarelli2010free}
Caffarelli, L.~A. and McCann, R.~J.
\newblock Free boundaries in optimal transport and {Monge-Ampere} obstacle problems.
\newblock \emph{Ann. Math.}, 171\penalty0 (2):\penalty0 673--730, Mar. 2010.

\bibitem[Cao et~al.(2018)Cao, Ma, Long, and Wang]{cao2018partial}
Cao, Z., Ma, L., Long, M., and Wang, J.
\newblock Partial adversarial domain adaptation.
\newblock In \emph{Proc. Eur. Conf. Comput. Vis. (ECCV)}, Munich, Germany, Sep. 2018.

\bibitem[Cao et~al.(2019)Cao, You, Long, Wang, and Yang]{cao2019learning}
Cao, Z., You, K., Long, M., Wang, J., and Yang, Q.
\newblock Learning to transfer examples for partial domain adaptation.
\newblock In \emph{Proc. IEEE/CVF Conf. Comput. Vis. Pattern Recognit. (CVPR)}, Long Beach, CA, USA, June 2019.

\bibitem[Cao et~al.(2022)Cao, You, Zhang, Wang, and Long]{9736609}
Cao, Z., You, K., Zhang, Z., Wang, J., and Long, M.
\newblock From big to small: Adaptive learning to partial-set domains.
\newblock \emph{IEEE Trans. Pattern Anal. Mach. Intell. (PAMI)}, 45\penalty0 (2):\penalty0 1766--1780, Mar. 2022.

\bibitem[Catoni(2007)]{catoni2007pac}
Catoni, O.
\newblock \emph{{PAC}-{Bayesian} Supervised Classification: the Thermodynamics of Statistical Learning}.
\newblock IMS Lecture Notes Monogr. Ser., 56, Beachwood, OH, \update{USA}, 2007.

\bibitem[Chang et~al.(2022)Chang, Shi, Tuan, and Wang]{chang2022unified}
Chang, W., Shi, Y., Tuan, H., and Wang, J.
\newblock Unified optimal transport framework for universal domain adaptation.
\newblock In \emph{Proc. Conf. Neural Inf. Process. Syst. (NeurIPS)}, New Orleans, \update{Louisiana,} USA, Nov. 2022.

\bibitem[Courty et~al.(2014)Courty, Flamary, and Tuia]{courty-14a}
Courty, N., Flamary, R., and Tuia, D.
\newblock Domain adaptation with regularized optimal transport.
\newblock In \emph{Proc. Mach. Learn. Knowl. Discov. Databases (ECML PKDD)}, Nancy, France, Sep. 2014.

\bibitem[Courty et~al.(2017{\natexlab{a}})Courty, Flamary, Habrard, and Rakotomamonjy]{courty2017joint}
Courty, N., Flamary, R., Habrard, A., and Rakotomamonjy, A.
\newblock Joint distribution optimal transportation for domain adaptation.
\newblock In \emph{Proc. Conf. Neural Inf. Process. Syst. (NeurIPS)}, Long Beach, CA, USA, Dec. 2017{\natexlab{a}}.

\bibitem[Courty et~al.(2017{\natexlab{b}})Courty, Flamary, Tuia, and Rakotomamonjy]{courty-17a}
Courty, N., Flamary, R., Tuia, D., and Rakotomamonjy, A.
\newblock Optimal transport for domain adaptation.
\newblock \emph{IEEE Trans. Pattern Anal. Mach. Intell. (PAMI)}, 39\penalty0 (9):\penalty0 1853--1865, Oct. 2017{\natexlab{b}}.

\bibitem[Damodaran et~al.(2018)Damodaran, Kellenberger, Flamary, Tuia, and Courty]{damodaran2018deepjdot}
Damodaran, B.~B., Kellenberger, B., Flamary, R., Tuia, D., and Courty, N.
\newblock {DeepJDOT}: Deep joint distribution optimal transport for unsupervised domain adaptation.
\newblock In \emph{Proc. Eur. Conf. Comput. Vis. (ECCV)}, Munich, Germany, Sep. 2018.

\bibitem[Farahani et~al.(2021)Farahani, Voghoei, Rasheed, and Arabnia]{farahani2021brief}
Farahani, A., Voghoei, S., Rasheed, K., and Arabnia, H.~R.
\newblock A brief review of domain adaptation.
\newblock In \emph{Adv. Data Sci. Inf. Eng.}, Springer, Cham, \update{Switzerland}, Oct. 2021.

\bibitem[Fatras et~al.(2021)Fatras, S{\'e}journ{\'e}, Flamary, and Courty]{fatras2021unbalanced}
Fatras, K., S{\'e}journ{\'e}, T., Flamary, R., and Courty, N.
\newblock Unbalanced minibatch optimal transport; applications to domain adaptation.
\newblock In \emph{Proc. Int. Conf. Mach. Learning (ICML)}, Virtual Conference, July 2021.

\bibitem[Figalli(2010)]{figalli2010optimal}
Figalli, A.
\newblock The optimal partial transport problem.
\newblock \emph{Arch. Ration. Mech. Anal.}, 195\penalty0 (2):\penalty0 533--560, Jan. 2010.

\bibitem[Flamary et~al.(2021)Flamary, Courty, Gramfort, Alaya, Boisbunon, Chambon, Chapel, Corenflos, Fatras, Fournier, Gautheron, Gayraud, Janati, Rakotomamonjy, Redko, Rolet, Schutz, Seguy, Sutherland, Tavenard, Tong, and Vayer]{flamary2021pot}
Flamary, R., Courty, N., Gramfort, A., Alaya, M.~Z., Boisbunon, A., Chambon, S., Chapel, L., Corenflos, A., Fatras, K., Fournier, N., Gautheron, L., Gayraud, N.~T., Janati, H., Rakotomamonjy, A., Redko, I., Rolet, A., Schutz, A., Seguy, V., Sutherland, D.~J., Tavenard, R., Tong, A., and Vayer, T.
\newblock {POT}: Python optimal transport.
\newblock \emph{J. Mach. Learn. Res. (JMLR)}, 22\penalty0 (78):\penalty0 1--8, Jan. 2021.

\bibitem[Ganin et~al.(2016)Ganin, Ustinova, Ajakan, Germain, Larochelle, Laviolette, Marchand, and Lempitsky]{ganin2016domain}
Ganin, Y., Ustinova, E., Ajakan, H., Germain, P., Larochelle, H., Laviolette, F., Marchand, M., and Lempitsky, V.
\newblock Domain-adversarial training of neural networks.
\newblock \emph{J. Mach. Learn. Res. (JMLR)}, 17\penalty0 (1):\penalty0 2096--2030, Jan. 2016.

\bibitem[Griffin et~al.(2022)Griffin, Holub, and Perona]{griffin2007caltech}
Griffin, G., Holub, A., and Perona, P.
\newblock Caltech 256, Apr. 2022.

\bibitem[Gu et~al.(2021)Gu, Yu, Sun, Xu, et~al.]{gu2021adversarial}
Gu, X., Yu, X., Sun, J., Xu, Z., et~al.
\newblock Adversarial reweighting for partial domain adaptation.
\newblock In \emph{Proc. Conf. Neural Inf. Process. Syst. (NeurIPS)}, Virtual Conference, Dec. 2021.

\bibitem[Gu et~al.(2024)Gu, Yu, Yang, Sun, and Xu]{Gu2024}
Gu, X., Yu, X., Yang, Y., Sun, J., and Xu, Z.
\newblock Adversarial reweighting with $\alpha$-power maximization for domain adaptation.
\newblock \emph{Int. J. Comput. Vis. (IJCV)}, 132\penalty0 (10):\penalty0 4768--4791, May 2024.

\bibitem[Guo et~al.(2022)Guo, Zhu, and Zhang]{guo2022selective}
Guo, P., Zhu, J., and Zhang, Y.
\newblock Selective partial domain adaptation.
\newblock In \emph{Proc. Br. Mach. Vis. Conf. (BMVC)}, London, UK, Nov. 2022.

\bibitem[He et~al.(2023)He, Li, Xia, Tang, Yang, and Ye]{he2023addressing}
He, C., Li, X., Xia, Y., Tang, J., Yang, J., and Ye, Z.
\newblock Addressing the overfitting in partial domain adaptation with self-training and contrastive learning.
\newblock \emph{IEEE Trans. Circuits Syst. Video Technol.}, 34\penalty0 (3):\penalty0 1532--1545, July 2023.

\bibitem[He et~al.(2016)He, Zhang, Ren, and Sun]{He_2016_CVPR}
He, K., Zhang, X., Ren, S., and Sun, J.
\newblock Deep residual learning for image recognition.
\newblock In \emph{Proc. IEEE/CVF Conf. Comput. Vis. Pattern Recognit. (CVPR)}, Las Vegas, USA, June 2016.

\bibitem[Hellström et~al.(2025)Hellström, Durisi, Guedj, and Raginsky]{hellstrom-25a}
Hellström, F., Durisi, G., Guedj, B., and Raginsky, M.
\newblock Generalization bounds: Perspectives from information theory and {PAC-Bayes}.
\newblock \emph{Found. Trends Mach. Learn.}, 18\penalty0 (1):\penalty0 1--223, Jan. 2025.

\bibitem[Li et~al.(2020)Li, Liu, Lin, Wen, Su, Huang, and Ding]{li2020deep}
Li, S., Liu, C.~H., Lin, Q., Wen, Q., Su, L., Huang, G., and Ding, Z.
\newblock Deep residual correction network for partial domain adaptation.
\newblock \emph{IEEE Trans. Pattern Anal. Mach. Intell. (PAMI)}, 43\penalty0 (7):\penalty0 2329--2344, Jan. 2020.

\bibitem[Li \& Chen(2022)Li and Chen]{li2022partial}
Li, W. and Chen, S.
\newblock Partial domain adaptation without domain alignment.
\newblock \emph{IEEE Trans. Pattern Anal. Mach. Intell. (PAMI)}, 45\penalty0 (7):\penalty0 8787--8797, Dec. 2022.

\bibitem[Liang et~al.(2020)Liang, Wang, Hu, He, and Feng]{liang2020balanced}
Liang, J., Wang, Y., Hu, D., He, R., and Feng, J.
\newblock A balanced and uncertainty-aware approach for partial domain adaptation.
\newblock In \emph{Proc. Eur. Conf. Comput. Vis. (ECCV)}, Virtual Conference, Aug. 2020.

\bibitem[Liang et~al.(2021)Liang, Hu, Wang, He, and Feng]{9512429}
Liang, J., Hu, D., Wang, Y., He, R., and Feng, J.
\newblock Source data-absent unsupervised domain adaptation through hypothesis transfer and labeling transfer.
\newblock \emph{IEEE Trans. Pattern Anal. Mach. Intell. (PAMI)}, 44\penalty0 (11):\penalty0 8602--8617, Aug. 2021.

\bibitem[Lin et~al.(2022)Lin, Zhou, Qiu, and Zheng]{lin2022adversarial}
Lin, K.-Y., Zhou, J., Qiu, Y., and Zheng, W.-S.
\newblock Adversarial partial domain adaptation by cycle inconsistency.
\newblock In \emph{Proc. Eur. Conf. Comput. Vis. (ECCV)}, Tel Aviv, Israel, Oct. 2022.

\bibitem[Lipton et~al.(2018)Lipton, Wang, and Smola]{lipton2018detecting}
Lipton, Z., Wang, Y.-X., and Smola, A.
\newblock Detecting and correcting for label shift with black box predictors.
\newblock In \emph{Proc. Int. Conf. Mach. Learning (ICML)}, Stockholm, Sweden, July 2018.

\bibitem[Long et~al.(2015)Long, Cao, Wang, and Jordan]{long2015learning}
Long, M., Cao, Y., Wang, J., and Jordan, M.
\newblock Learning transferable features with deep adaptation networks.
\newblock In \emph{Proc. Int. Conf. Mach. Learning (ICML)}, Lille, France, July 2015.

\bibitem[Long et~al.(2018)Long, Cao, Wang, and Jordan]{NIPS2018_7436}
Long, M., Cao, Z., Wang, J., and Jordan, M.~I.
\newblock Conditional adversarial domain adaptation.
\newblock In \emph{Proc. Conf. Neural Inf. Process. Syst. (NeurIPS)}, Montréal, Canada, Dec. 2018.

\bibitem[Luo \& Ren(2023)Luo and Ren]{Luo_2023_CVPR}
Luo, Y.-W. and Ren, C.-X.
\newblock {MOT}: Masked optimal transport for partial domain adaptation.
\newblock In \emph{Proc. IEEE/CVF Conf. Comput. Vis. Pattern Recognit. (CVPR)}, Vancouver, Canada, June 2023.

\bibitem[Luo \& Ren(2024)Luo and Ren]{luo2024invariant}
Luo, Y.-W. and Ren, C.-X.
\newblock When invariant representation learning meets label shift: Insufficiency and theoretical insights.
\newblock \emph{IEEE Trans. Pattern Anal. Mach. Intell. (PAMI)}, 46\penalty0 (12):\penalty0 9407--9422, June 2024.

\bibitem[McAllester(1999)]{mcallester1999pac}
McAllester, D.~A.
\newblock {PAC-Bayesian} model averaging.
\newblock In \emph{Proc. Conf. Comput. Learn. Theory (COLT)}, Santa Cruz, CA, USA, July 1999.

\bibitem[Nagarajan \& Kolter(2019)Nagarajan and Kolter]{nagarajan2019uniform}
Nagarajan, V. and Kolter, J.~Z.
\newblock Uniform convergence may be unable to explain generalization in deep learning.
\newblock In \emph{Proc. Conf. Neural Inf. Process. Syst. (NeurIPS)}, Vancouver, Canada, Dec. 2019.

\bibitem[Nguyen et~al.(2022)Nguyen, Nguyen, Pham, Ho, et~al.]{nguyen2022improving}
Nguyen, K., Nguyen, D., Pham, T., Ho, N., et~al.
\newblock Improving mini-batch optimal transport via partial transportation.
\newblock In \emph{Proc. Int. Conf. Mach. Learning (ICML)}, Baltimore, \update{MA}, USA, July 2022.

\bibitem[Ohnishi \& Honorio(2021)Ohnishi and Honorio]{ohnishi2021novel}
Ohnishi, Y. and Honorio, J.
\newblock Novel change of measure inequalities with applications to {PAC-Bayesian} bounds and {Monte Carlo} estimation.
\newblock In \emph{Proc. Int. Conf. Artif. Intell. Stat. (AISTATS)}, Virtual Conference, Apr. 2021.

\bibitem[Panareda~Busto \& Gall(2017)Panareda~Busto and Gall]{panareda-busto17-10a}
Panareda~Busto, P. and Gall, J.
\newblock Open set domain adaptation.
\newblock In \emph{Proc. IEEE Int. Conf. Comput. Vis. (ICCV)}, Venice, Italy, Oct. 2017.

\bibitem[Redko et~al.(2017)Redko, Habrard, and Sebban]{redko2017theoretical}
Redko, I., Habrard, A., and Sebban, M.
\newblock Theoretical analysis of domain adaptation with optimal transport.
\newblock In \emph{Proc. Mach. Learn. Knowl. Discov. Databases (ECML PKDD)}, Skopje, Macedonia, Sep. 2017.

\bibitem[Redko et~al.(2019)Redko, Morvant, Habrard, Sebban, and Bennani]{redko2020survey}
Redko, I., Morvant, E., Habrard, A., Sebban, M., and Bennani, Y.
\newblock \emph{Advances in domain adaptation theory}.
\newblock Elsevier, Oxford, UK, 2019.

\bibitem[Russakovsky et~al.(2015)Russakovsky, Deng, Su, Krause, Satheesh, Ma, Huang, Karpathy, Khosla, Bernstein, et~al.]{russakovsky2015imagenet}
Russakovsky, O., Deng, J., Su, H., Krause, J., Satheesh, S., Ma, S., Huang, Z., Karpathy, A., Khosla, A., Bernstein, M., et~al.
\newblock {ImageNet} large scale visual recognition challenge.
\newblock \emph{Int. J. Comput. Vis. (IJCV)}, 115\penalty0 (3):\penalty0 211--252, Apr. 2015.

\bibitem[Sahoo et~al.(2023)Sahoo, Panda, Feris, Saenko, and Das]{sahoo2021select}
Sahoo, A., Panda, R., Feris, R., Saenko, K., and Das, A.
\newblock Select, label, and mix: Learning discriminative invariant feature representations for partial domain adaptation.
\newblock In \emph{Proc. IEEE/CVF Winter Conf. Appl. Comput. Vis.}, \update{Waikoloa,} Hawaii, USA, Jan. 2023.

\bibitem[Shen et~al.(2018)Shen, Qu, Zhang, and Yu]{shen2018wasserstein}
Shen, J., Qu, Y., Zhang, W., and Yu, Y.
\newblock Wasserstein distance guided representation learning for domain adaptation.
\newblock In \emph{Proc. AAAI Conf. Artif. Intell. (AAAI)}, New Orleans, \update{Louisiana,} USA, Apr. 2018.

\bibitem[Tachet~des Combes et~al.(2020)Tachet~des Combes, Zhao, Wang, and Gordon]{tachet2020domain}
Tachet~des Combes, R., Zhao, H., Wang, Y.-X., and Gordon, G.~J.
\newblock Domain adaptation with conditional distribution matching and generalized label shift.
\newblock In \emph{Proc. Conf. Neural Inf. Process. Syst. (NeurIPS)}, Virtual Conference, Dec. 2020.

\bibitem[Tzeng et~al.(2017)Tzeng, Hoffman, Saenko, and Darrell]{tzeng2017adversarial}
Tzeng, E., Hoffman, J., Saenko, K., and Darrell, T.
\newblock Adversarial discriminative domain adaptation.
\newblock In \emph{Proc. IEEE/CVF Conf. Comput. Vis. Pattern Recognit. (CVPR)}, \update{Honolulu,} Hawaii, USA, July 2017.

\bibitem[Venkateswara et~al.(2017)Venkateswara, Eusebio, Chakraborty, and Panchanathan]{office-home}
Venkateswara, H., Eusebio, J., Chakraborty, S., and Panchanathan, S.
\newblock Deep hashing network for unsupervised domain adaptation.
\newblock In \emph{Proc. IEEE/CVF Conf. Comput. Vis. Pattern Recognit. (CVPR)}, \update{Honolulu,} Hawaii, USA, July 2017.

\bibitem[Wainwright(2019)]{wainwright-19a}
Wainwright, M.~J.
\newblock \emph{High-Dimensional Statistics: a Non-Asymptotic Viewpoint}.
\newblock Cambridge Univ. Press, Cambridge, U.K., 2019.

\bibitem[Wang et~al.(2022)Wang, Xue, Lei, and Xia]{Wang2022PartialWA}
Wang, Z.-M., Xue, N., Lei, L., and Xia, G.-S.
\newblock Partial {Wasserstein} adversarial network for non-rigid point set registration.
\newblock In \emph{Proc. Int. Conf. Learn. Represent. (ICLR)}, Virtual Conference, Apr. 2022.

\bibitem[Wang et~al.(2024)Wang, Xue, Lei, J{\"o}rnsten, and Xia]{wang2024partial}
Wang, Z.-M., Xue, N., Lei, L., J{\"o}rnsten, R., and Xia, G.-S.
\newblock Partial distribution matching via partial {Wasserstein} adversarial networks.
\newblock \emph{arXiv}, Sep. 2024.

\bibitem[Wu et~al.(2023)Wu, Wu, Chen, Jin, Cui, Cao, and Li]{9957101}
Wu, K., Wu, M., Chen, Z., Jin, R., Cui, W., Cao, Z., and Li, X.
\newblock Reinforced adaptation network for partial domain adaptation.
\newblock \emph{IEEE Trans. Circuits Syst. Video Technol.}, 33\penalty0 (5):\penalty0 2370--2380, Nov. 2023.

\bibitem[Xiao et~al.(2021)Xiao, Ding, and Liu]{xiao2021implicit}
Xiao, W., Ding, Z., and Liu, H.
\newblock Implicit semantic response alignment for partial domain adaptation.
\newblock In \emph{Proc. Conf. Neural Inf. Process. Syst. (NeurIPS)}, Virtual Conference, Dec. 2021.

\bibitem[Yang et~al.(2023)Yang, Cheung, Ding, Tan, Xue, and Zhang]{9705553}
Yang, C., Cheung, Y.-M., Ding, J., Tan, K.~C., Xue, B., and Zhang, M.
\newblock Contrastive learning assisted-alignment for partial domain adaptation.
\newblock \emph{IEEE Trans. Neural Netw. Learn. Syst.}, 34\penalty0 (10):\penalty0 7621--7634, Feb. 2023.

\bibitem[Zhang et~al.(2021)Zhang, Bengio, Hardt, Recht, and Vinyals]{zhang-21a-rethinking}
Zhang, C., Bengio, S., Hardt, M., Recht, B., and Vinyals, O.
\newblock Understanding deep learning (still) requires rethinking generalization.
\newblock \emph{Commun. ACM}, 64\penalty0 (3):\penalty0 107--115, Feb. 2021.

\bibitem[Zhang et~al.(2018)Zhang, Ding, Li, and Ogunbona]{zhang2018importance}
Zhang, J., Ding, Z., Li, W., and Ogunbona, P.
\newblock Importance weighted adversarial nets for partial domain adaptation.
\newblock In \emph{Proc. IEEE/CVF Conf. Comput. Vis. Pattern Recognit. (CVPR)}, Salt Lake City, UT, USA, June 2018.

\end{thebibliography}
